\documentclass[10pt,twocolumn,letterpaper]{article}
\usepackage[pagenumbers]{cvpr}
\usepackage{graphicx}
\usepackage{amsmath}
\usepackage{amssymb}
\usepackage{booktabs}
\usepackage{paralist}
\usepackage{blindtext}
\usepackage{titling}
\usepackage{tabularx} 
\usepackage{multirow}
\usepackage[pagebackref,breaklinks,colorlinks,citecolor=blue]{hyperref}
\usepackage{comment}
\usepackage{grffile}
\usepackage[capitalize]{cleveref}
\crefname{section}{Sec.}{Secs.}
\Crefname{section}{Section}{Sections}
\Crefname{table}{Table}{Tables}
\crefname{table}{Tab.}{Tabs.}
\usepackage{xcolor}

\newcommand{\clip}[0]{\textrm{CLIP}}

\newcommand{\R}{\mathbb R}

\newlength{\figsep}
\begin{document}

\title{Solving Inverse Problems with NerfGANs}

\author{Giannis Daras$^\textrm{\textdagger}$\thanks{This work was done during an internship at Google Research.} \\ {\tt\small giannisdaras@utexas.edu} \and
Wen-Sheng Chu$^\ddagger$ \\ {\tt\small wschu@google.com} \and
Abhishek Kumar$^\ddagger$ \\ {\tt\small abhishk@google.com} \and
Dmitry Lagun$^\ddagger$ \\ {\tt\small dlagun@google.com} \and
Alexandros G. Dimakis$^\textrm{\textdagger}$ \\ {\tt\small dimakis@austin.utexas.edu}
}
\date{$^\textrm{\textdagger}$University of Texas at Austin \hspace{8mm} $^\ddagger$Google Research}

\maketitle

\begin{abstract}
We introduce a novel framework for solving inverse problems using NeRF-style generative models. We are interested in the problem of 3-D scene reconstruction given a single 2-D image and known camera parameters. We show that naively optimizing the latent space leads to artifacts and poor novel view rendering. 
We attribute this problem to volume obstructions that are clear in the 3-D geometry and become visible in the renderings of novel views.
We propose a novel radiance field regularization method to obtain better 3-D surfaces and improved novel views given single view observations. Our method naturally extends to general inverse problems including inpainting where one observes only partially a single view.
We experimentally evaluate our method, achieving visual improvements and performance boosts over the baselines in a wide range of tasks.  Our method achieves $30-40\%$ MSE reduction and $15-25\%$ reduction in LPIPS loss compared to the previous state of the art. 
\end{abstract}

\begin{figure}[t!]
    \centering
    \includegraphics[width=\linewidth]{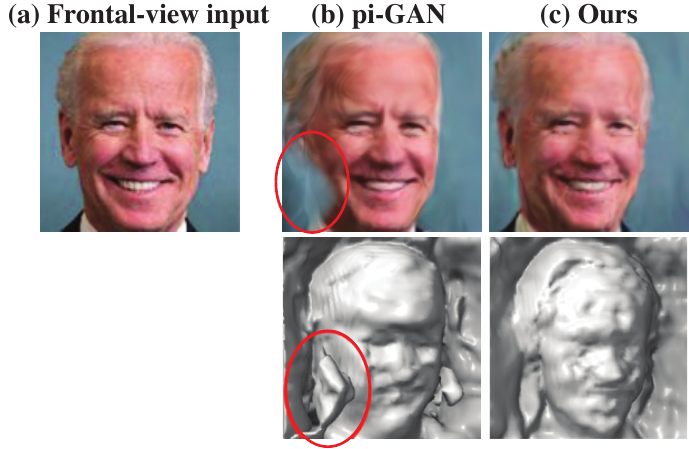}
    \caption{
    {\bf NerfGAN inversion:}
    Given a single frontal view image, we would like to generate novel angle views and the underlying 3-D geometry. 
    As shown, latent space optimization as proposed in pi-GAN~\cite{chan2021pi} creates obstructions (\textit{stones}) that can produce artifacts in novel views. 
    Middle column illustrates the artifacts in the rendered image (top row) and the 3-D geometry (bottom row). 
    Our inversion algorithm removes these issues by optimizing over both the 2-D view and the 3-D shape of the radiance field. 
    Middle Column: Reconstructed 3-D geometry and novel view using pi-GAN direct latent space optimization. 
    Right column: Generated 3-D structure and novel view using the proposed reconstruction algorithm. 
    We emphasize that our algorithm also uses the same generator (pi-GAN), but recovers a better latent vector compared to direct latent-space optimization. 
    This leads to a better 3-D geometry reconstruction and better 2-D novel views. 
    }
    \vspace{\figsep}
    \label{fig:fig1}
\end{figure}

\section{Introduction}

State-of-the-art generative models have become capable of generating extremely high-fidelity images of the 2-D world~\cite{biggan, ylg, esser2021taming, stylegan, stylegan2}.  Despite their wide success, current generative models often fail to capture the 3-D structure of the represented scenes and offer limited control over the geometrical properties of the generated images.

NerfGANs~\cite{meng2021gnerf, nguyen2019hologan, niemeyer2021giraffe, chan2021pi} are a new family of generative models that directly model the 3-D space by leveraging the success of Neural Radiance Fields (NeRFs)~\cite{nerf}. NerfGANs generate 3-D structure in the form of a Radiance Field and then output 2-D images by rendering the field from different camera views. They are not yet as competitive as state-of-the-art 2-D models for image generation~\cite{stylegan, stylegan2, stylegan3, biggan, ylg, daras2020smyrf}. However, modeling directly the 3-D space offers many new possibilities, beyond generating photo-realistic images, that have yet to be explored.

We study how we can use pre-trained NerfGANs as a prior to solve inverse problems.
We start with the problem of single-view inversion: given a single 2-D image (\eg, a photograph of a person) and known camera parameters we want to create novel views and reconstruct the 3-D geometry leveraging a pre-trained NerfGAN, e.g. the 
state of the art pi-GAN~\cite{chan2021pi}.
We denote with $G(z, p)$ a 3-D NerfGAN that takes a latent vector $z \in \mathbb{R}^k$ and a 3-D space position $p$ in $\mathbb{R}^3$ and outputs a color and a density value. For a given latent vector $z$, the NerfGAN scene can be rendered as a 2-D image for any camera position. 
Formally, for a given camera position $c$ the produced 2-D image is denoted by $\mathcal{R}(c, G(z, \cdot)) $, where $\mathcal{R}$ is the rendering operator. 
Given a single target image $x^*$ and known camera parameters $c$, NerfGAN inversion is the problem of finding the optimal latent code $z^*$ that creates a 3-D scene that renders to $x^*$. 

A natural method for NerfGAN inversion (used in pi-GAN~\cite{chan2021pi}) is to optimize the latent vector to match the observed target image~\cite{bora2017compressed}: 
\begin{gather}
\min_z || \mathcal{R}(c, G(z, \cdot )) - x^* ||. 
\label{eq:csgm}
\end{gather}
We unveil a major limitation of this vanilla method. 
The produced neural radiance field renders correctly to the given target image $x^*$, but even small rotations produce significant artifacts in novel views. 
In Figure \ref{fig:fig1}, a single front view was given and a radiance field was reconstructed using latent space optimization. When rendering a novel side view, a blue shade artifact appears in the bottom left as annotated.  We discover that neural fields created by solving the inverse problem often have three dimensional obstructions (we call them \textit{stones}) that are invisible in the frontal view but create visual artifacts in the renderings of novel views.

\begin{figure}[!tp]
    \begin{subfigure}[t]{0.15\textwidth}
            \begin{center}
                \includegraphics[width=\linewidth]{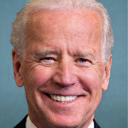}
                \caption{Reference}
            \end{center}
    \end{subfigure} \hspace{1mm}
    \begin{subfigure}[t]{0.15\textwidth}
            \begin{center}
                \includegraphics[width=\linewidth]{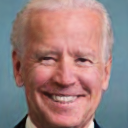}
                \caption{Ours (MSE: 0.0045, LPIPS: 44.18)}
            \end{center}
    \end{subfigure} \hspace{1mm} 
    \begin{subfigure}[t]{0.15\textwidth}
            \begin{center}
                \includegraphics[width=\linewidth]{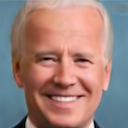}
                \caption{pi-GAN (MSE: 0.0078, LPIPS: 221.07)}
            \end{center}
    \end{subfigure} \\
    \begin{subfigure}[t]{0.15\textwidth}
            \begin{center}
                \includegraphics[width=\linewidth]{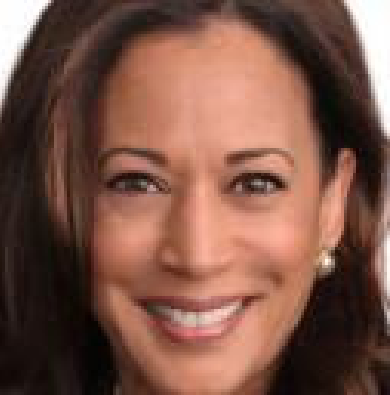} 
                \caption{Reference \\}
            \end{center}
    \end{subfigure} \hspace{1mm} 
    \begin{subfigure}[t]{0.15\textwidth}
            \begin{center}
                \includegraphics[width=\linewidth]{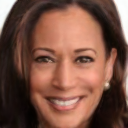} 
                \caption{Ours (MSE: 0.0029, LPIPS: 28.11)}
            \end{center}
    \end{subfigure} \hspace{1mm} 
    \begin{subfigure}[t]{0.15\textwidth}
            \begin{center}
                \includegraphics[width=\linewidth]{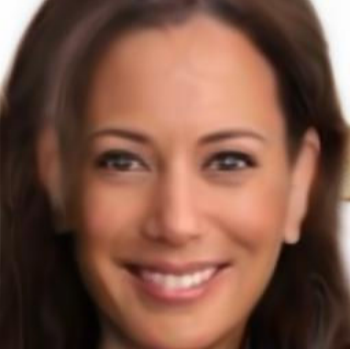}
                \caption{pi-GAN (MSE: 0.0060, LPIPS: 28.90)}
            \end{center}
    \end{subfigure}
    \caption{\textbf{Regularization impact on frontal-view reconstruction:}
    Given the frontal-view images (a), (d), we show reconstructions with our method in (b), (e) and pi-GAN in (e), (f). Our method leverages a more expressive 3-D consistency loss, yielding better MSE and LPIPS results than pi-GAN that uses distance from average frequencies.
    Note that the pi-GAN images are directly taken from the original paper \cite{chan2021pi}.
    }
    \label{fig:biden_kamala_frontal}
    \vspace{\figsep}
\end{figure}

Similar issues were raised in the pi-GAN~\cite{chan2021pi} paper which observed hollow-face artifacts in generated images but not obstructions. The authors of~\cite{chan2021pi} identify this as an open problem: \textit{
``In certain cases, pi-GAN can generate a radiance field that creates viable images when rendered from each direction but nonetheless fails to conform to the 3-D shape that we would expect. Further investigation may reveal insights that could resolve such ambiguities.''}. We observe that solving inverse problems using direct latent space optimization, as in \eqref{eq:csgm}, frequently produces unrealistic 3-D obstructions that also lead to visual artifacts when rendered from novel views. To account for this problem, the authors of~\cite{chan2021pi} proposed to penalize divergence between the SIREN~\cite{siren} frequencies and phase shifts and their average values. We show that this approach, even though it produces smooth geometries, significantly reduces the range of the generator, leading to blurred reconstructions (see Figure \ref{fig:biden_kamala_frontal}).
Instead, our method solves for
\begin{gather}
\min_z || \mathcal{R}(c, G(z, \cdot )) - x^* ||+ \lambda S_{3-D}(  G(z, \cdot )),
\label{eq:csgmreg}
\end{gather}
where $S_{3-D}$ imposes a high penalty for latent vectors $z$ that create unnatural geometries. As we explain subsequently, we achieve this 3-D regularization by creating a convex combination of distances to reference geometries.

\noindent \textbf{Extension to General Inverse problems:} 
Beyond reconstructing the 3-D structure of a scene using a single view $x^*$, we extend our method to general inverse problems. For example, our method can be directly applied when there are missing pixels in the view (inpainting), a blurred observed view, or observations of the single view with random projections 
or Fourier projections arising in medical imaging and compressed sensing~\cite{bora2017compressed,jalal2021mri}. 

Consider the general setting where the unknown image is $x^*$ and we observe 
\begin{align}
    y = \mathcal{A}[x^*] + \eta, 
\end{align}
where $\mathcal{A}$ is the forward operator that somehow corrupts the image (e.g. pixel removal, blurring, or projections) and $\eta$ is a noise vector.

Direct latent optimization in this case would correspond to finding the latent vector $z$ that best explains the measurements, as expressed below:
\begin{gather}
\min_z || \mathcal{A} [\mathcal{R}(c, G(z, \cdot )] - y ||. 
\label{eq:gen}
\end{gather}
This natural baseline creates 3-D obstructions (thus artifacts) as in single view inversion. 
We propose to solve the optimization problem with the same 3-D regularization: 
\begin{gather}
    \min_z ||  \mathcal{A} [ \mathcal{R}(c, G(z, \cdot ))] - y ||+ \lambda S_{\text{3D}}(  G(z, \cdot )).
    \label{eq:gensol}
\end{gather}
Our method can be applied to linear inverse problems (where $\mathcal{A}$ is a matrix) or even non-linear such as phase retrieval~\cite{hand2018phase} as long as the $\mathcal A$ is differentiable almost everywhere. 

The key issue is to devise a 3-D regularizer, $S_{3-D}$, that does not lead to measurements overfitting, i.e. big real error $|| \mathcal{R}(c, G(z, \cdot )) - x||$ with small measurements error (first term of \eqref{eq:gensol}). We use a set of reference geometries and an annealing mechanism in gradient descent to lock-in on better fitting geometries as we subsequently explain.

\begin{figure*}[t!]
    \centering
    \includegraphics[width=\linewidth]{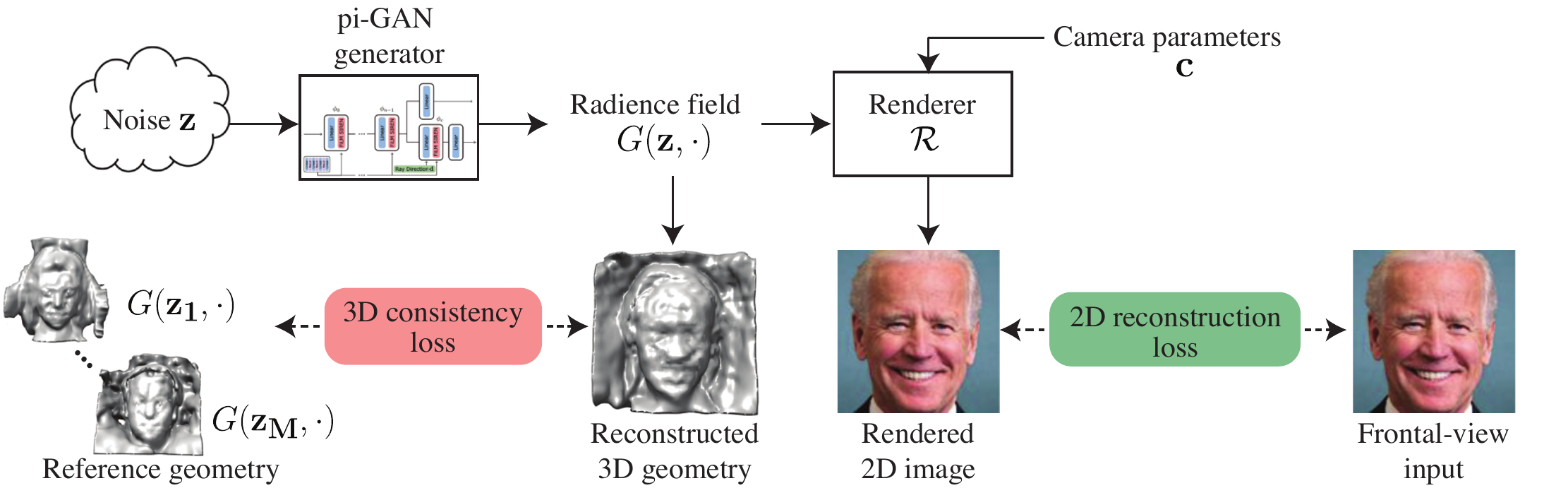}
    \caption{
    {\bf An overview of the proposed method:}
    We jointly optimize over 2-D re-construction of the input image and 3-D consistency of the radiance field.
    Given a randomly sampled noise vector $\textbf{z}$ from a Gaussian distribution, we obtain radiance field $G(\mathbf{z},\cdot)$ via the pi-GAN generator $G$, and rendered 2-D images using a conventional volumetric renderer $\mathcal{R}$ for given camera parameters $\mathbf{c}$. Our loss minimizes at the same time the distance of the rendered image to the given view and a 3-D consistency term that penalizes unrealistic geometries.
    }
    \vspace{\figsep}
    \label{fig:fig2}
\end{figure*}

To the best of our knowledge, this paper shows for the first time the challenges involved in using NerfGANs for solving inverse problems and proposes a principled framework to address them.  

\noindent \textbf{Our Contributions:}
\begin{compactitem}
    \item We identify a key limitation on reconstructing 3-D geometries and novel views from a single reference image.
    We demonstrate that existing algorithms are not sufficient, because either: i) they overfit to the measurements and produce artifacts to novel views (Figure \ref{fig:fig1}), or ii) they significantly limit the expressive power of the generator (Figure \ref{fig:biden_kamala_frontal}).
    \item We trace the problem back to unrealistic 3-D geometries that need to be avoided in the course of the optimization. We propose a principled framework for regularizing the radiance field itself, without limiting the range of the generator. Our framework drives the network to generate realistic geometries by penalizing distance from a set of realistic geometries under a novel 3-D loss.
    \item We show how to obtain a candidates set of realistic geometries using CLIP~\cite{clip} and a pre-trained NerfGAN.
    \item We experimentally evaluate our method achieving visual improvements and performance boosts over the baselines in a wide range of tasks. Our method achieves $30-40\%$ MSE and $15-25\%$ reduction in LPIPS~\cite{lpips} compared to the previous state of the art. 
\end{compactitem}

\section{Related Work}

\paragraph{NerfGANs}
NerfGANs are NeRF~\cite{nerf} style generators that are trained adversarially~\cite{goodfellow2014generative}. Recently, there has been tremendous progress in making NeRFs more efficient \cite{liu2020neural,Lindell20arxiv_AutoInt,Rebain20arxiv_derf,Garbin21arxiv_FastNeRF,reiser2021kilonerf, neff2021donerf} and  extending them beyond static scenes \cite{Li20arxiv_nsff,Xian20arxiv_stnif,Du20arxiv_nerflow,li2021neural}. As larger and more realistic 3-D Neural radiance field generators are made possible, we expect that solving inverse problems with them to become increasingly relevant for numerous applications. 

In this paper, we investigate the challenges in solving inverse problems with NerfGANs that are stemming from unrealistic 3-D structures.
Our work can in principle be used with any generator from the growing family of NerfGANs~\cite{chan2021pi, niemeyer2021giraffe, schwarz2021graf, nguyen2019hologan, meng2021gnerf}. For the purposes of this paper, we use the pi-GAN~\cite{chan2021pi} generator since, to the best of our knowledge, there is no other NerfGAN paper that discusses inverse problems. A concurrent work~\cite{pan2021shadingguided} proposes architectural changes to the generator to obtain smoother geometries. Potentially, our method could yield even better results with this new generator, but this remains as future work.

A new line of research~\cite{zhou2021cips3d, gu2021stylenerf} uses a small NerfGAN to produce a coarse image representation at low-resolution and then a 2-D network to transform it to photorealistic, higher-resolution image. These approaches side-step the computational cost of training a traditional NerfGAN at high resolutions and present promising results in solving inverse problems. These techniques are orthogonal to our method for obtaining better 3-D geometries from single image-views and can be combined as a post-processing step to improve the quality of our generated views.

\begin{figure*}[t!]
    \centering
    \includegraphics[width=\linewidth]{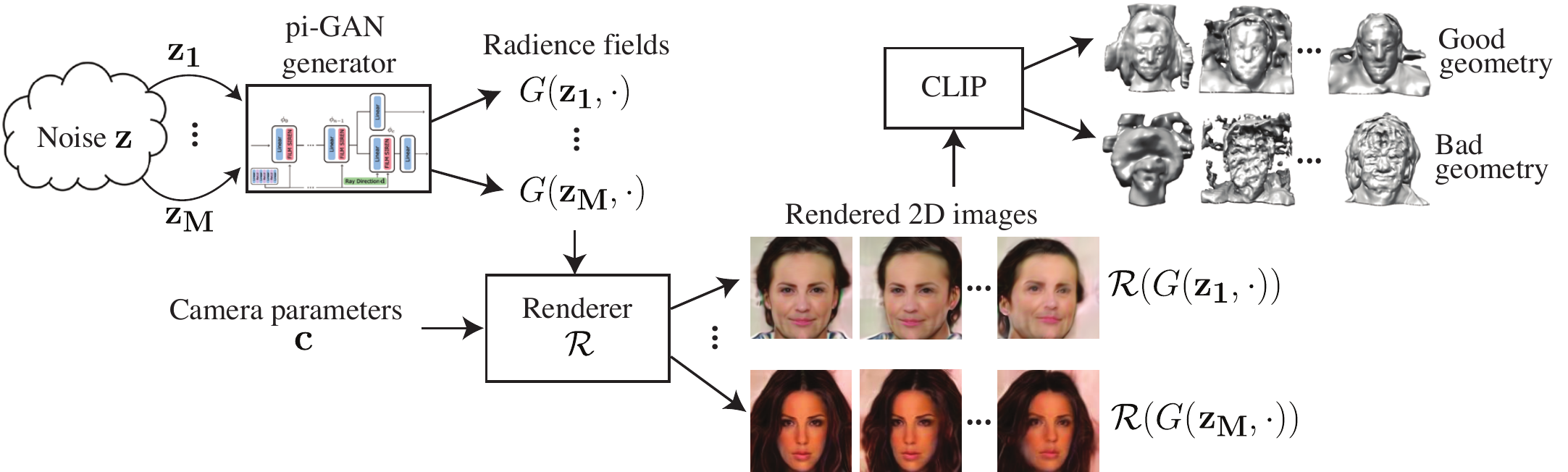}
    \caption{
    {\bf Candidate selection for reference geometries:} We use CLIP~\cite{clip}, a pre-trained free-text image classifier, to identify whether a radiance field is realistic and to form a set of reference geometries. The idea is that for bad geometries, certain attributes (such as CLIP's belief on how realistic an image is) should change, even for small changes in the camera parameters. Given rendered 2-D images from randomly sampled noise vectors $\{\mathbf{z_1},\dots,\mathbf{z_M}\}$, we observe differences in CLIP's logits to identify realistic geometries.}
    \label{fig:clip_preprocessing}
    \vspace{\figsep}
\end{figure*}

\paragraph{Inverse Problems} Our method relies on significant prior work in unsupervised methods for inverse problems using pre-trained generative models.
We note that our method leverages a pre-trained generator, hence avoiding the computational overhead of training an end-to-end supervised network to go from observations to 3-D geometry~\cite{nguyenphuoc2018rendernet, NEURIPS2018_92cc2275}. There are various benefits of using unsupervised methods for solving inverse problems including robustness to data structure shifts and unknown variations in the corruption process~\cite{ongie2020deep}. Compared to classical approaches for 3-D reconstruction from a single view, like the seminal 3DMM algorithm~\cite{3dmm}, our method is more general since it can be applied for any corrupted image, as long as the forward operator that corrupts the image is known and differentiable.

Inversion algorithms are building on latent space optimization as proposed in the CSGM~\cite{bora2017compressed} algorithm and we extend it using Perceptual Loss ~\cite{lpips, ilo} and Geodesic regularization for frequencies, inspired by StyleGAN2 and PULSE~\cite{stylegan, stylegan2, pulse}. 
Our central innovation is the 3-D regularization term which is directly applied in the neural radiance field and is, to the best of knowledge, entirely novel. 

\section{Method}

\paragraph{Motivation}

Figure \ref{fig:biden_kamala_frontal}(c), \ref{fig:biden_kamala_frontal}(f), shows inversions that appear in the pi-GAN paper. The first thing to notice is that the reconstructed images are sufficiently different compared to the input images. The poor frontal view reconstruction could be attributed to the limited range of the pi-GAN generator. Figure \ref{fig:biden_kamala_frontal}(b), \ref{fig:biden_kamala_frontal}(e) shows that this is not true; using the exact same generator we are able to get much better reconstructions compared to pi-GAN as shown visually and by the MSE and Perceptual losses. 

The pi-GAN reconstructions look like smoothed versions of the reference images. This superficial smoothness comes from the regularization term that is used in the pi-GAN paper. Specifically, the authors of~\cite{chan2021pi} penalize divergence between the SIREN~\cite{siren} frequencies and phase shifts and their average values. Figure \ref{fig:fig1}(b) shows a novel view obtained with pi-GAN without this regularization. The face is much closer to the given input. There is a caveat though; the novel view shows artifacts (circled with red color). To understand where these artifacts are coming from, in Figure \ref{fig:fig1}(b), second row, we visualize the 3-D geometries that correspond to the obtained frequencies and phase shifts. We see that there are three dimensional obstructions (we call them stones) that are invisible in the frontal view. This experiment shows an important trade-off, i.e. matching better the measurements but with poor generalization vs inferior reconstruction of the image but smoother novel views.

A natural question is why the optimization trajectory finds such geometries that match almost perfectly the given 2-D image but produce artifacts in novel views. The short answer is that these unrealistic geometries are fairly common, even in pure image generation with Gaussian inputs. We perform an experiment to validate this.

For this experiment, we are using CLIP~\cite{clip}, a free text image classifier. CLIP can take as input an image and a text, and it produces text and image embeddings that have big inner product if the text describes well what is portrayed in the image. The idea of the experiment is that for bad geometries, certain attributes of the object change as you slightly move the camera, so that should be reflected in the CLIP logits. CLIP has been used before as a cherry-picking tool for image generation~\cite{dalle}, but we show a natural extension for identifying realistic 3-D structure.

We first collect latent vectors, sampled from a Gaussian distribution, and form the set $\mathcal S = \{z_1, ..., z_M\}$. 
For each of the elements in the set, we render images from camera positions $c_1, ..., c_K$, and form the set $\mathcal S_z = \{\mathcal R(c_1, G(z, \cdot)), ..., \mathcal R(c_K, G(z, \cdot))\}$. 
Each of these sets is assigned a cost given by the maximum difference of CLIP logits between the images in the set and the text prompt $T = $\texttt{"A non-corrupted, non-noisy image of a person."}. 
We use this prompt because we expect that for side views, bad geometries will show more artifacts due to 3-D obstructions (stones) outside of the face.

The cost $w_z$ of $\mathcal S_z$ is given by:

\begin{gather}
    w_z = \max_{s_1, s_2 \in \mathcal S_z} \left|\clip(s_1, T) - \clip(s_2, T)\right|.
    \label{eq:consistency_cost}
\end{gather}

After assigning the costs, the set of latents that correspond to unrealistic geometries is given by:
\begin{gather}
    \textrm{Bad}_{z, \epsilon} = \{z \in S | w_z \geq \epsilon \}.
\end{gather}

The fraction of unrealistic geometries, \ie, $\frac{|\textrm{Bad}_{z, \epsilon}|}{|S|},$
is a measure of how often do bad geometries occur in the range of a NerfGAN. We found that such geometries are common in the range of pi-GAN~\cite{chan2021pi} -- approximately $40\%$ of geometries are classified as ``bad" by CLIP. 
The top-right portion of Figure \ref{fig:clip_preprocessing} illustrates geometries classified as bad by CLIP. 
More examples of bad geometries, their corresponding 2-D views and more details are provide in the Appendix.

\paragraph{Obtaining realistic reference geometries}
Similarly, one can use CLIP to identify realistic geometries. For reasons that will be explained in Section \ref{sec:optimization}, we are interested in collecting a set of realistic geometries that render to visually plausible 2-D images. For each Gaussian sampled $z$, we assign two costs: i) the \textit{consistency cost} we defined in \eqref{eq:consistency_cost} and ii) the plausibility cost:
\begin{gather}
    c_z = \min_{s \in \mathcal S_z} \clip(s, T).
\end{gather}
We finally collect the set:
\begin{gather}
    \textrm{Good}_{z, \epsilon_1, \epsilon_2} = \{z \in S | w_z \leq \epsilon_1, c_z \leq \epsilon_2 \}.
\end{gather}
The whole procedure is illustrated in Figure \ref{fig:clip_preprocessing}, with examples of geometries and renderings.

\paragraph{Optimization problem}
\label{sec:optimization}
Without regularization, the optimization trajectory often reaches points of minimum loss but with poor generalization to novel views. A natural idea is to regularize towards realistic geometries. We will use the notation $G_\sigma(z, \cdot)$ to denote the density part of the radiance field and $P \in \mathbb R^{k \times 3}$ to denote its discrete representation.

A straightforward way to constrain the 3-D shape is to force it to be close to a 3-D geometry that is known to be good. We collect a set of latent vectors $S$ that correspond to realistic geometries $\{G_\sigma(z, \cdot) | z \in S \}$ and try to regularize the inferred geometry towards the \emph{most suitable} geometry in our realistic set. This can be made more concrete in the form of the following optimization problem:

\begin{align}
\begin{split}
    \min_{z \in \mathbb R^d}  \frac{1}{2}||\mathcal R(c, G_\sigma(z, \cdot)) - x^*||^2 \\ \textrm{s.t.} \ \min_{z_{\textrm{ref}} \in \mathcal S} \mathcal L\left(G_\sigma(z, P), G_\sigma(z_{\textrm{ref}}, P)\right) \leq \epsilon.
    \label{eq:min_min_with_const}
\end{split}
\end{align}

To solve this problem with Gradient Descent, we write down the penalized version and solve for:
\begin{gather}
    \min_{z \in \mathbb R^d, z_{\textrm{ref}} \in \mathcal S}  \frac{1}{2}||\mathcal R(c, G(z, \cdot)) - x^*||^2 \nonumber \\ 
    + \lambda \cdot \mathcal L\left(G_{\sigma}(z, P ), G_{\sigma}(z_{\textrm{ref}}, P )\right).
    \label{eq:min_min}
\end{gather}

There are two issues with the formulation of \eqref{eq:min_min}. First, it is a min-min problem where the inner minimum is over a discrete set. Gradient Descent (GD) is likely to get stuck to a local minima: the reference geometry that happens to be closer to the initialization is likely to be the active constraint of \eqref{eq:min_min_with_const}, even though it might not be the one that minimizes the total objective. We observe this problem experimentally, for more see Figure \ref{fig:anplot}. The second issue is that the two terms in our loss function might be incompatible. For example, if the reference set $\mathcal S$ is small, there might be no geometry that renders to the measurements.

\begin{figure}
    \centering
    \raisebox{8pt}{\includegraphics[width=0.32\linewidth]{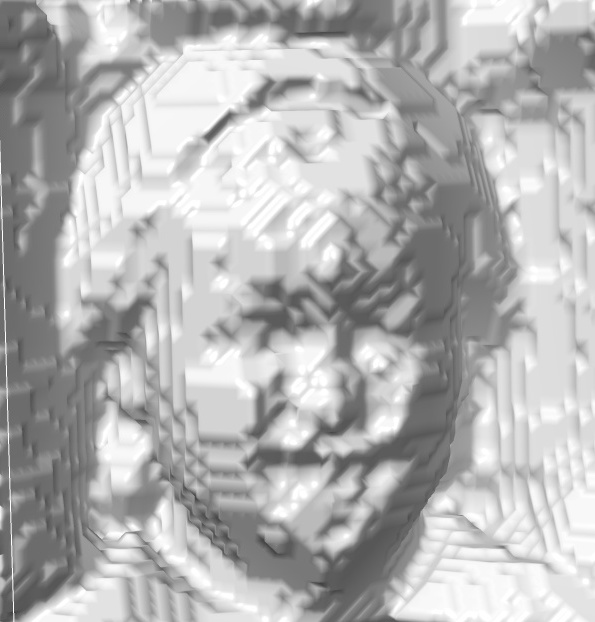}} \;
    \includegraphics[width=0.32\linewidth]{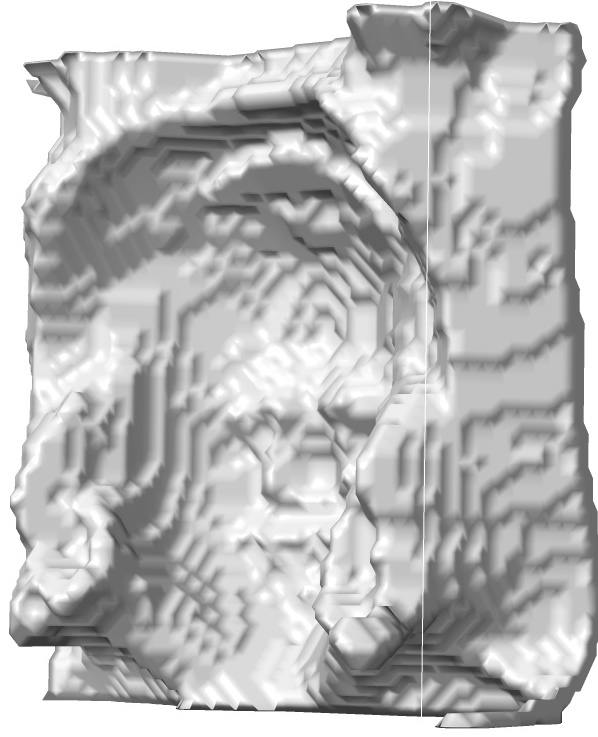}
    \caption{
    {\bf Reference geometry:}
    Illustration of a reference geometry (front and back) extracted with the Marching Cubes algorithm used in our set of reference geometries $\{G(z_i, \cdot)\}$.}
    \label{fig:facial_mask}
    \vspace{\figsep}
\end{figure}

We propose a relaxation of the objective of \eqref{eq:min_min}, where the $\min$ is replaced with a soft-operator that allows all reference geometries to contribute to the gradients based on how close they are, under $\mathcal L$, to the current radiance field. 
We consider the following optimization problem:
\begin{align}
\begin{split}
    \min_{z} \frac{1}{2}||\mathcal R(c, G(z, \cdot)) - x^* ||^2 + \\ + \lambda \sum_{z_{\textrm{ref}} \in S} \left(\frac{e^{ -\delta  \mathcal L_{z, z_{\textrm{ref}}} }}{\sum_{z_{\textrm{ref}}' \in S} e^{ -\delta  \mathcal L_{z, z_{\textrm{ref}'}}}}\right) \mathcal L_{z, z_{\textrm{ref}}}.
    \label{eq:tau}
\end{split}
\end{align}
This can be interpreted as finding a nonnegative weighting of the losses $\mathcal{L}_{z,z_{\textrm{ref}}}$ that solves $\min_{w\geq 0} \sum_i w_i \mathcal{L}_{z,z_{{\textrm{ref}}_i}}$, such that $D(u,w)\leq\gamma(\delta)$ for some divergence measure $D$, uniform distribution $u$, and a radius parameter $\gamma$. 
The softmax weighting emerges when $D$ is taken to be KL-divergence \cite{kumar2021constrained}.
Observe that for $\delta \to \infty$ (corresponding to a large enough $\gamma$), the optimization problems of Equations \eqref{eq:min_min}, \eqref{eq:tau} have the same solution. However, this formulation is more powerful since it allows blending of the reference geometries in case we cannot match the measurements otherwise. If the distribution of the contributions of each loss is close to a Dirac, then we have 3-D consistency.

In all our experiments, we use GD to solve the optimization problem of \eqref{eq:tau}. We gradually anneal the temperature parameter $\delta$ during the course of the optimization to encourage convergence to a single reference geometry. We choose the $z$ that minimizes the total loss. If the loss curve is flattened, we prefer the $z$ that corresponds to lower temperature because it correlates with better 3-D consistency.

\paragraph{Choice of 3-D loss function} 
Figure \ref{fig:fig1} shows that when solving inverse problems without 3-D regularization, the reconstructed 3-D geometries have what we call ``stones", i.e. regions of high density outside of the face that create artifacts when we render the field with different camera parameters. To regularize for the stones, for each of the reference radiance fields, we obtain a face surface mask on the 3-D space and constrain the reconstructed voxel grid to match the reference outside of this 3-D mask. The intuition is that voxels outside of the facial surface should have low values (as in the reference geometries). Matching only these voxels gives our method enough flexibility to adjust the facial 3-D structure to match the measurements without having high density clusters (stones) outside of the face.

Formally, let $\mathcal M(p): \mathbb R^3 \to \{0, 1\}$ be the operator that gives the facial mask. We define our loss function as:

\begin{gather}
    \mathcal L_{z, z_{\textrm{ref}}} = ||\left(G_\sigma(z, P) - G_\sigma(z_{\textrm{ref}}, P)\right) \odot \left(1 - \mathcal M(P)\right)||_F
\end{gather}
where $||\cdot ||_F$ denotes the Frobenius norm.

In all our experiments, we use the vertices of the generated polygons of the Marching Cubes algorithm~\cite{lorensen1987marching} to get our facial mask. Figure \ref{fig:facial_mask} shows an example geometry and the corresponding mask for a latent in our reference set.

\begin{figure}
    \centering
    \includegraphics[width=\linewidth]{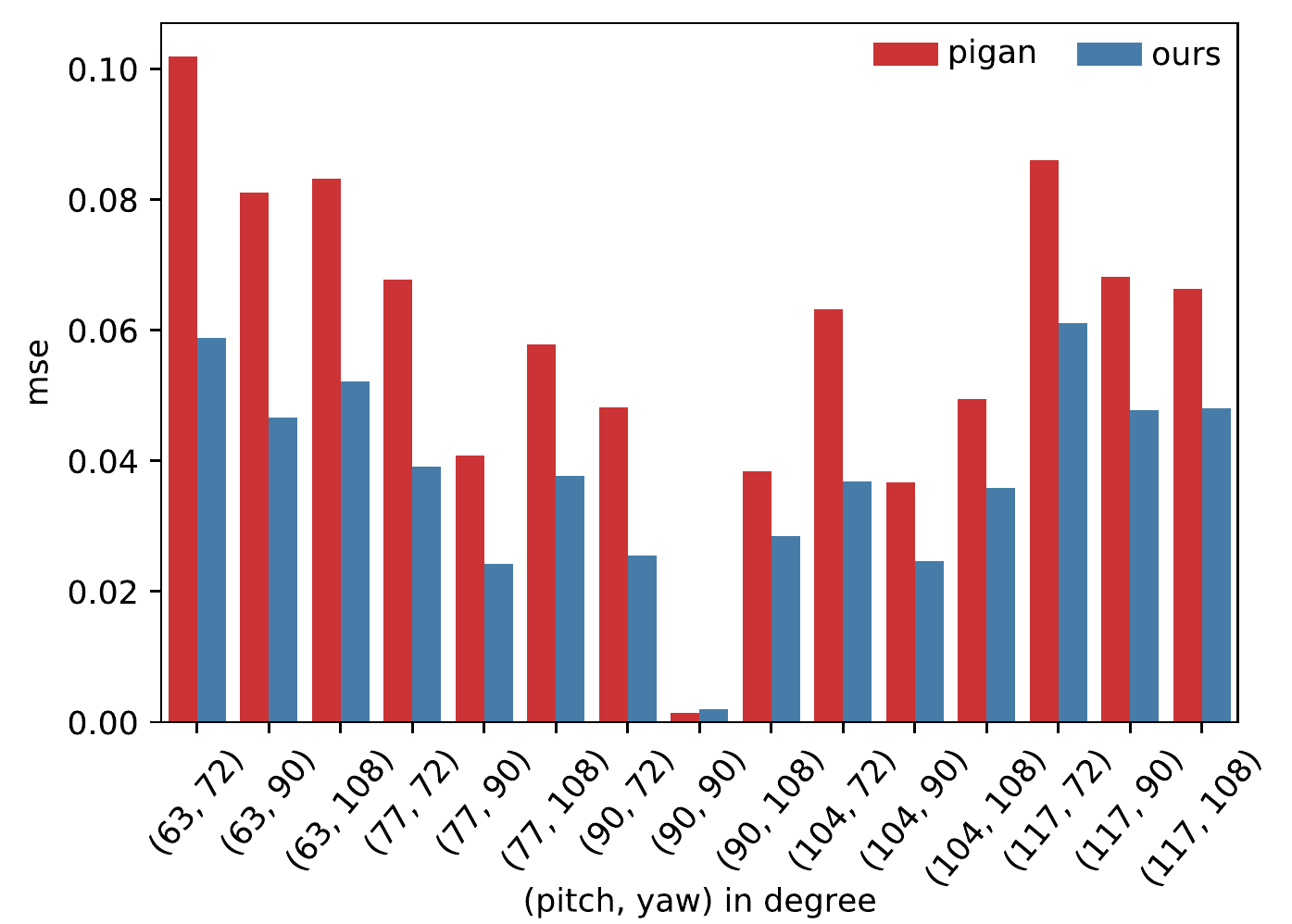}
    \includegraphics[width=\linewidth]{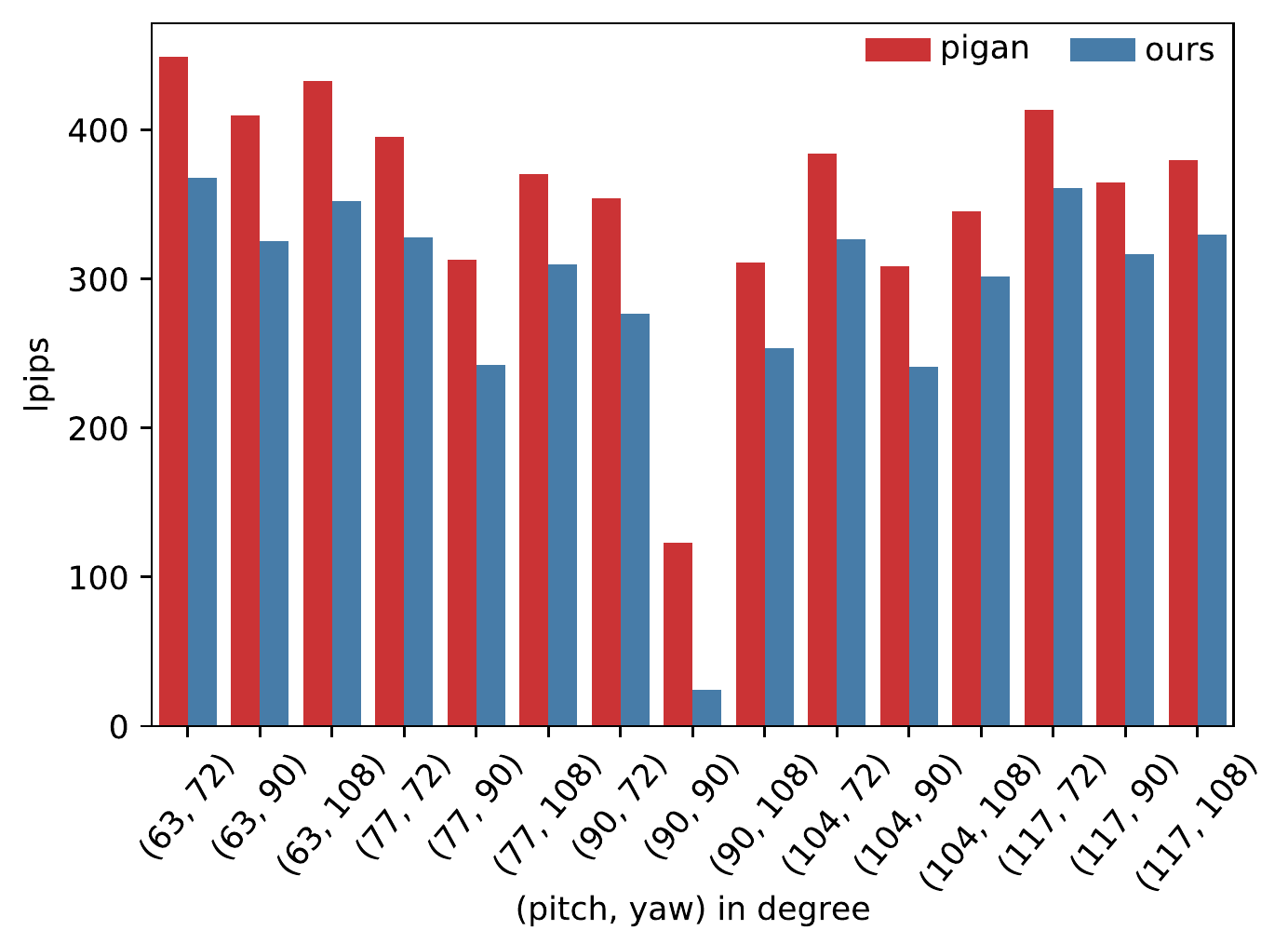}
    \caption{\textbf{MSE vs LPIPS:}
    Comparisons between pi-GAN \cite{chan2021pi} and our method on different camera angles on MSE and LPIPS metrics.
    MSE (top) indicates 2-D pixel reconstruction, where our method shows comparable loss in the given frontal view at (90,90), and noticeably lower loss in all other novel views.
    LPIPS (bottom) describes the perceptual loss, suggesting that our method has consistently less perceptual differences than pi-GAN.
    }
    \label{fig:mse}
    \vspace{\figsep}
\end{figure}

\section{Experimental Results}

In all our experiments, we use the pi-GAN~\cite{chan2021pi} generator, pre-trained on faces from CelebA~\cite{liu2018large}.
At each optimization step, our method generates a voxel grid using the current latent $z$. One natural question is how coarse the voxel grid representation should be in order for the regularization to be effective. We perform an extensive analysis on the Appendix. In short, our finding is that a voxel grid at resolution $32\times32\times32$ works well across all the considered tasks. We use a reference set of $64$ geometries, automatically selected with CLIP and manually inspected to ensure it has the desired quality and diversity. Our method runs in approximately $3$ minutes per image on a workstation of 4 V100 GPUs. For more experimental details, including all our hyperparameters, we refer the reader to the Appendix.

\subsection{Inversion} 
Our first comparison is with the regularization proposed in the pi-GAN paper ($\ell_2$ penalty for divergence from the average frequencies and phase shifts). For a fair comparison, we use the reconstructions directly from the pi-GAN paper. Figure \ref{fig:biden_kamala_frontal} shows the results for the frontal view. 
Our method significantly outperforms pi-GAN on the frontal view -- we observe $42\%$ reduction in MSE and $80\%$ reduction in LPIPS for the first image. As shown in the pi-GAN paper, their method indeed gives novel views without artifacts, but since the distance to the ground truth is very large, we do not consider it further in our experiments in the main paper. 
We refer the interested reader to the Appendix for a more extensive quantitative comparison with this method.

In the rest of the paper, we compare with the unregularized baseline that follows the CSGM~\cite{bora2017compressed} approach to match the measurements, i.e. solves the problem defined in \eqref{eq:csgm}. We run our method and the CSGM baseline on the images from the pi-GAN paper and for some images in the range of NerfGAN. For all our experiments, input is the frontal view (one can run our method for any view, as long as the camera parameters are known).

\begin{figure}[t!]
    \centering
    \newlength{\ww}\setlength{\ww}{0.222\linewidth}
    \begin{tabular}{@{}c@{\;\;}c@{}c@{}c@{}}
        \includegraphics[width=\ww]{images/biden.png} &
        \raisebox{20pt}{(a)}
        \includegraphics[width=\ww]{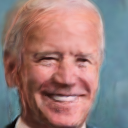} &
        \includegraphics[width=\ww]{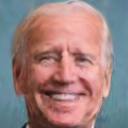} &
        \includegraphics[width=\ww]{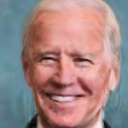} \\
        
        & \raisebox{20pt}{(b)} \includegraphics[width=\ww]{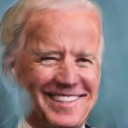} &
        \includegraphics[width=\ww]{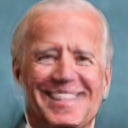} &
        \includegraphics[width=\ww]{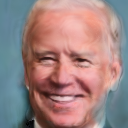} \\
        
        \includegraphics[width=\ww]{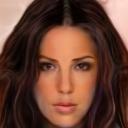} &
        \raisebox{20pt}{(c)} \includegraphics[width=\ww]{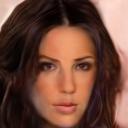} &
        \includegraphics[width=\ww]{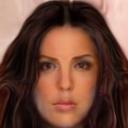} &
        \includegraphics[width=\ww]{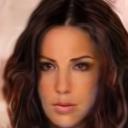} \\
        
        & \raisebox{20pt}{(d)} \includegraphics[width=\ww]{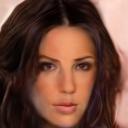} &
        \includegraphics[width=\ww]{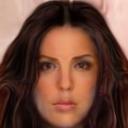} & 
        \includegraphics[width=\ww]{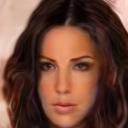} \\

        & \raisebox{20pt}{(e)} \includegraphics[width=\ww]{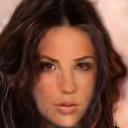} &
        \includegraphics[width=\ww]{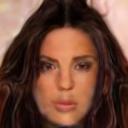} &
        \includegraphics[width=\ww]{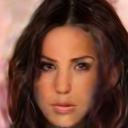} \\
    \end{tabular}
    \caption{
        {\bf Novel views}: Rows (a)(d) show novel views of our method,  and rows (b)(e) show the baseline. The input image, frontal view, is shown in the left column. The second input image is synthetic, so we also show ground truth novel views on Row (c). Our method removes artifacts that appear in the baseline, such as blurry blobs in Row (b) and unrealistic eyes in Row (e).}
    \label{fig:novel_views}
    \vspace{\figsep}
\end{figure}

Figure \ref{fig:novel_views} illustrates novel views rendered with our method and the baseline (for video results, see the Supplementary). For the images in the range, we have ground truth novel views that we also show in the Figure (obtained by inputting the corresponding viewing angle into the NerfGAN generator).
As shown, our method produces less artifacts (e.g. the blurry blobs shown in Row (b) is removed, together with the unrealistic artifacts in the eyes in Row (e). For the synthetic image, our method (Row (d)) almost matches exactly the ground truth novel views presented in Row (c). For the real image, we do not have ground truth novel views, and we do see some blurriness (even in our method) that can be attributed to the limited expressivity of the generator.

In Figure \ref{fig:mse} we show a quantitative comparison between our method and standard pi-GAN inversion for different views, measured in a set of $100$ synthetic images for which we have ground truth for all views. Our method achieves $30-40\%$ MSE reduction and $15-25\%$ reduction in LPIPS compared to latent space optimization without the 3-D regularization in the novel views. In the frontal view, the methods perform on par, suggesting overfitting of the baseline to the measurements.

\subsection{Inpainting} 
We now consider the problem of inpainting where one does not observe a full view $x^*$ but rather a known subset of pixels is missing.

In Figure \ref{fig:inpainting_plot} we plot the Mean Squared Error (MSE) versus the ratio of observed pixels for a novel view for the task of the randomized inpainting, i.e. a fraction of the pixels, selected at random, is missing each time. As shown, latent-space optimization baseline has an increasing MSE as the number of observations increases. This happens because the baseline is overfitting to the frontal view and fails to reconstruct the novel views correctly. 
In contrast, our method consistently produces lower MSE for all the novel views. 

Our MSE is almost constant in the considered range since our method gets optimal reconstruction in the considered range. For the frontal view, we observe that the MSE drops for both methods in the same way as the number of measurements increases. Our method gets $0.022$ MSE for $10\%$ inpainted pixels and $0.001$ for $100\%$ observed, while the baseline performs on par: $0.023$ observed for $10\%$ and $0.001$ for $100\%$ observed.

\subsection{Ablation Studies}

\begin{figure}
    \centering
    \begin{tabular}{@{}c@{}c@{}}
        \includegraphics[width=0.224\textwidth]{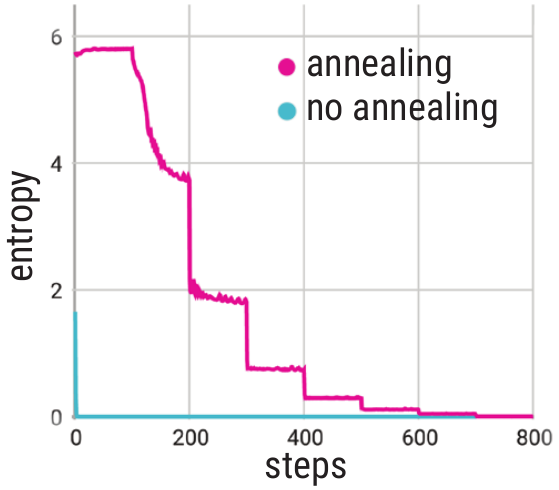} &
        \includegraphics[width=0.49\linewidth]{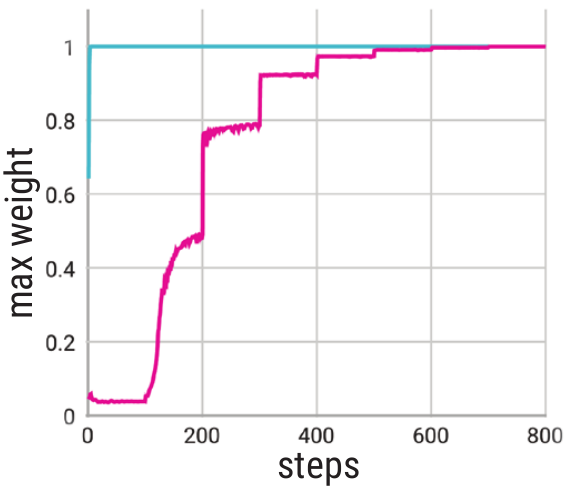} \\
        \includegraphics[width=0.49\linewidth]{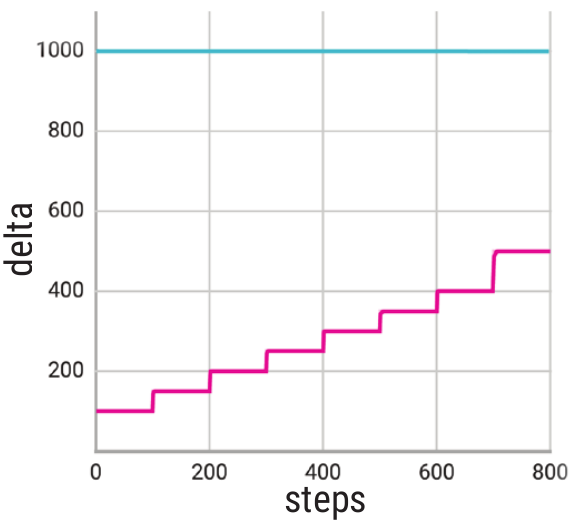} &
        \scriptsize
        \raisebox{45pt}{
            \begin{tabular}{@{\;}l@{\quad}c@{\quad}c@{\;}}
                \toprule
                & \multicolumn{2}{c}{MSE error} \\
                \cmidrule{2-3}
                (pitch, yaw) & No Annealing &  Annealing. \\
                \midrule
                (77, 90)  & 2.8e-2 & \textbf{2.2e-2} \\
                (90, 90)  & 5.5e-3 & \textbf{3.5e-3} \\
                (103, 90) & 3.2e-2 & \textbf{2.7e-2} \\
                (90, 99)  & 9.3e-3 & \textbf{5.4e-3} \\
                \bottomrule
            \end{tabular}
        }
    \end{tabular}
    \caption{\textbf{Temperature Annealing:} We ablate the annealing component of our method. Without annealing, the optimization locks immediately in one geometry (maximum weight of the blue line becomes $1$ after one step) and tries to minimize the measurements error. However, with annealing of $\delta$, we still converge to a single distribution (max weight of the purple line becomes $1$ eventually), but the model finds a geometry that gives better MSE scores consistently in all views, as illustrated in the Table. 
        }
    \label{fig:anplot}
\end{figure}

\paragraph{Temperature annealing} 
In the method section, we motivated the need to converge as much as possible to a single geometry. However, in the early stages of the optimization, we want to allow all reference geometries to contribute to the gradients since otherwise gradient descent might get trapped in a local minima. To achieve this, we do temperature annealing: in the early steps of gradient descent we have a high temperature (i.e. we allow our distribution over the references to look like uniform) and as we progress we decrease the temperature (increase $\delta$) to converge to a single radiance field. In all our experiments, we do step annealing, increasing $\delta$ by $50$ every $100$ optimization steps. Figure \ref{fig:anplot} shows how $\delta$, the entropy of the distribution over the references and its maximum weight evolve over time, with and without annealing. As shown, without annealing, the model converges to a single geometry immediately (maximum weight of the blue line becomes $1$ after one step), and won't change throughout the optimization. This leads to suboptimal results, as shown in the Table of Figure \ref{fig:anplot}. With annealing, in the early stages, the distribution has high entropy (close to uniform) and as time progresses we converge to a single radiance field. Our annealing scheme allows the modle to discover the truly best geometry to match the measurements.
\paragraph{Additional Experiments} In the Appendix, we present several additional experiments on real data as well as inverse problems including compressed sensing, super-resolution and inpainting. Additionally, inspired by the work of~\cite{zhou2021cips3d}, we show that we can project our renderings for novel views back to the range of a 2-D generator using Intermediate Layer Optimization~\cite{ilo} and achieve StyleGAN quality results for rendered views. We note that this is not part of our central innovation, but one can use it to get finer details in the rendered images. Our method can also generalize to objects beyond faces. For a relevant discussion and results, please refer to the Appendix.

\begin{figure}
    \centering
    \begin{tabular}{@{}c@{}c@{}}
        \includegraphics[width=0.5\linewidth]{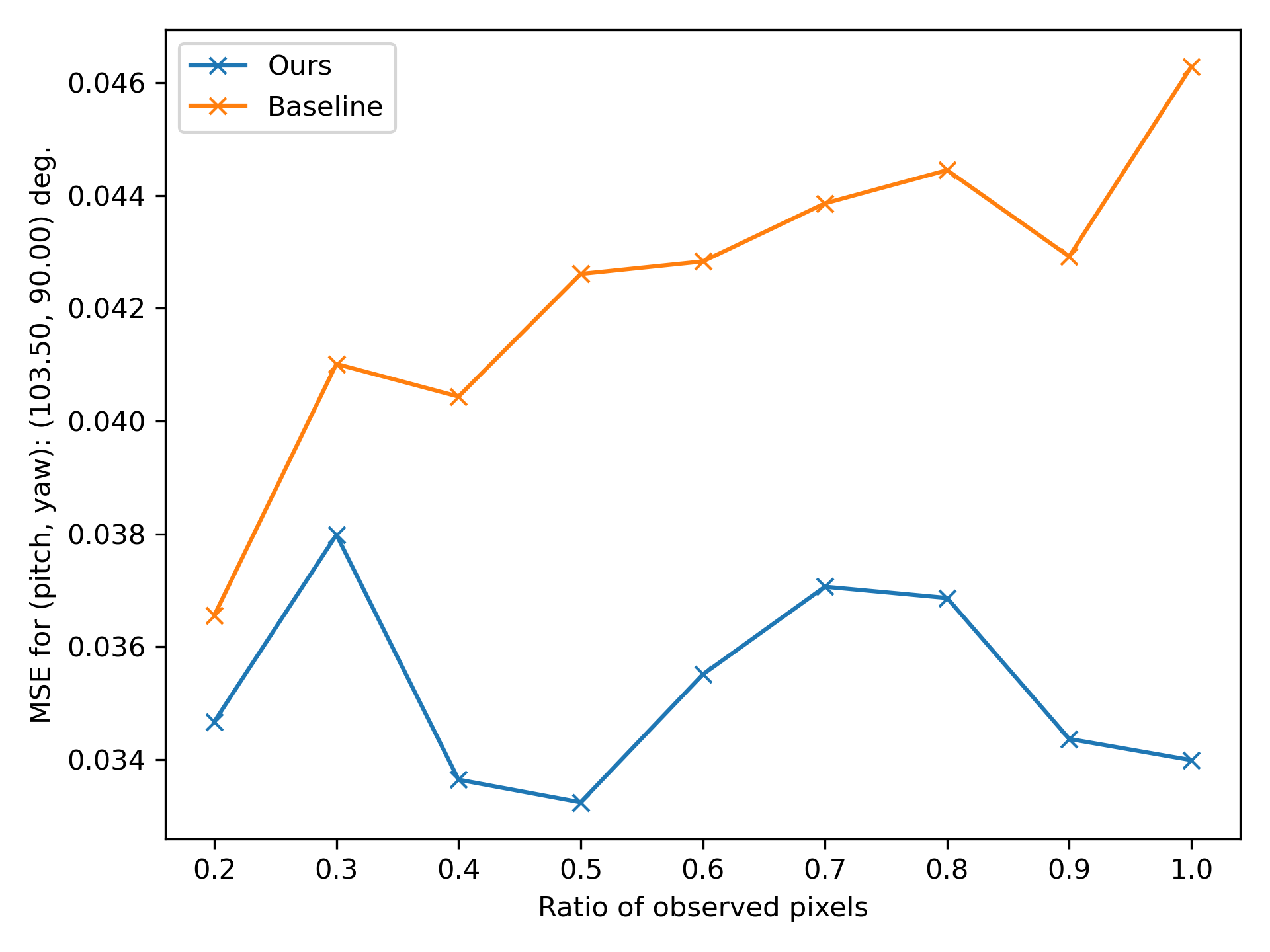} &
        \includegraphics[width=0.5\linewidth]{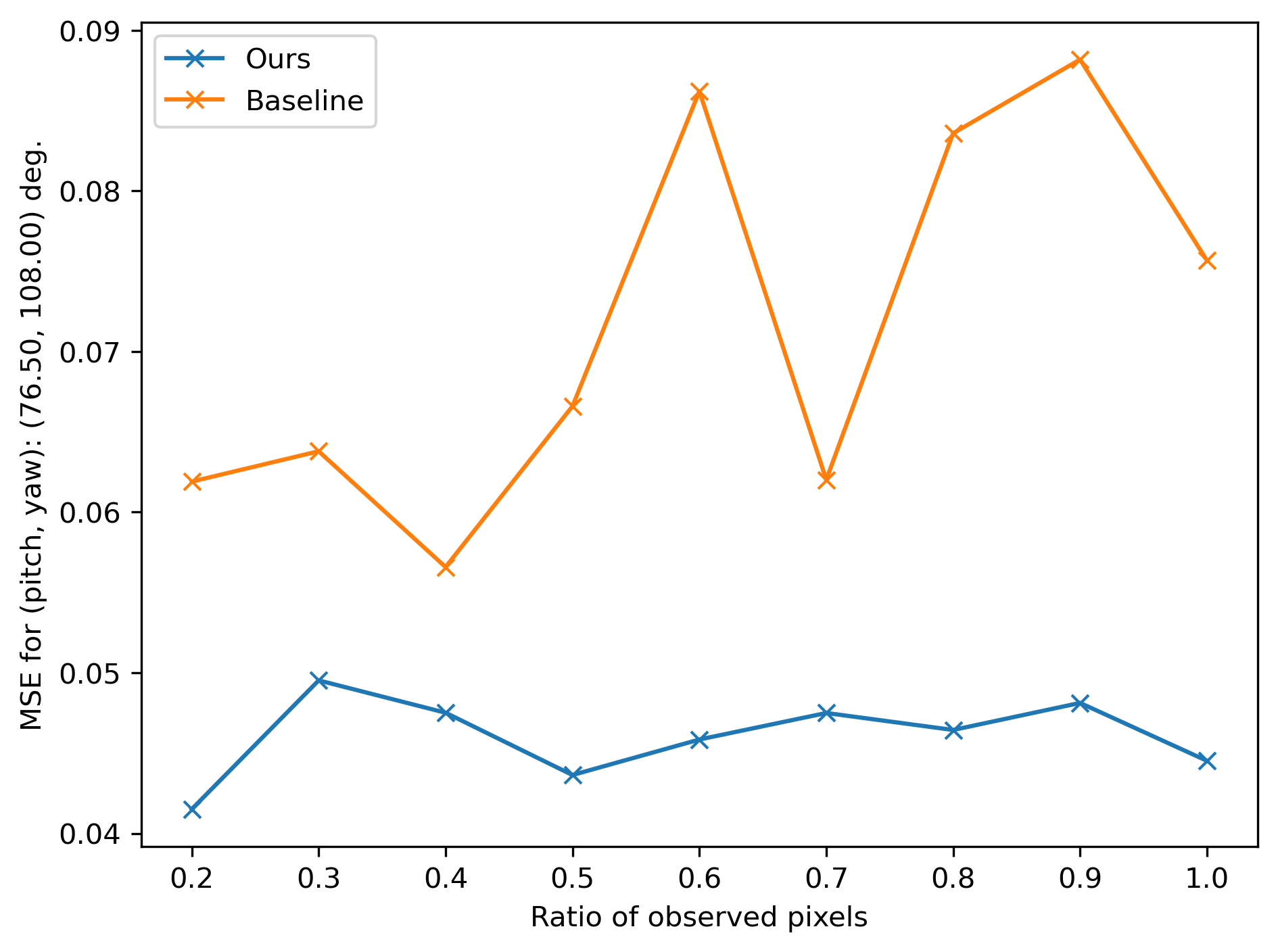} \\
        \centering
        \includegraphics[width=0.5\linewidth]{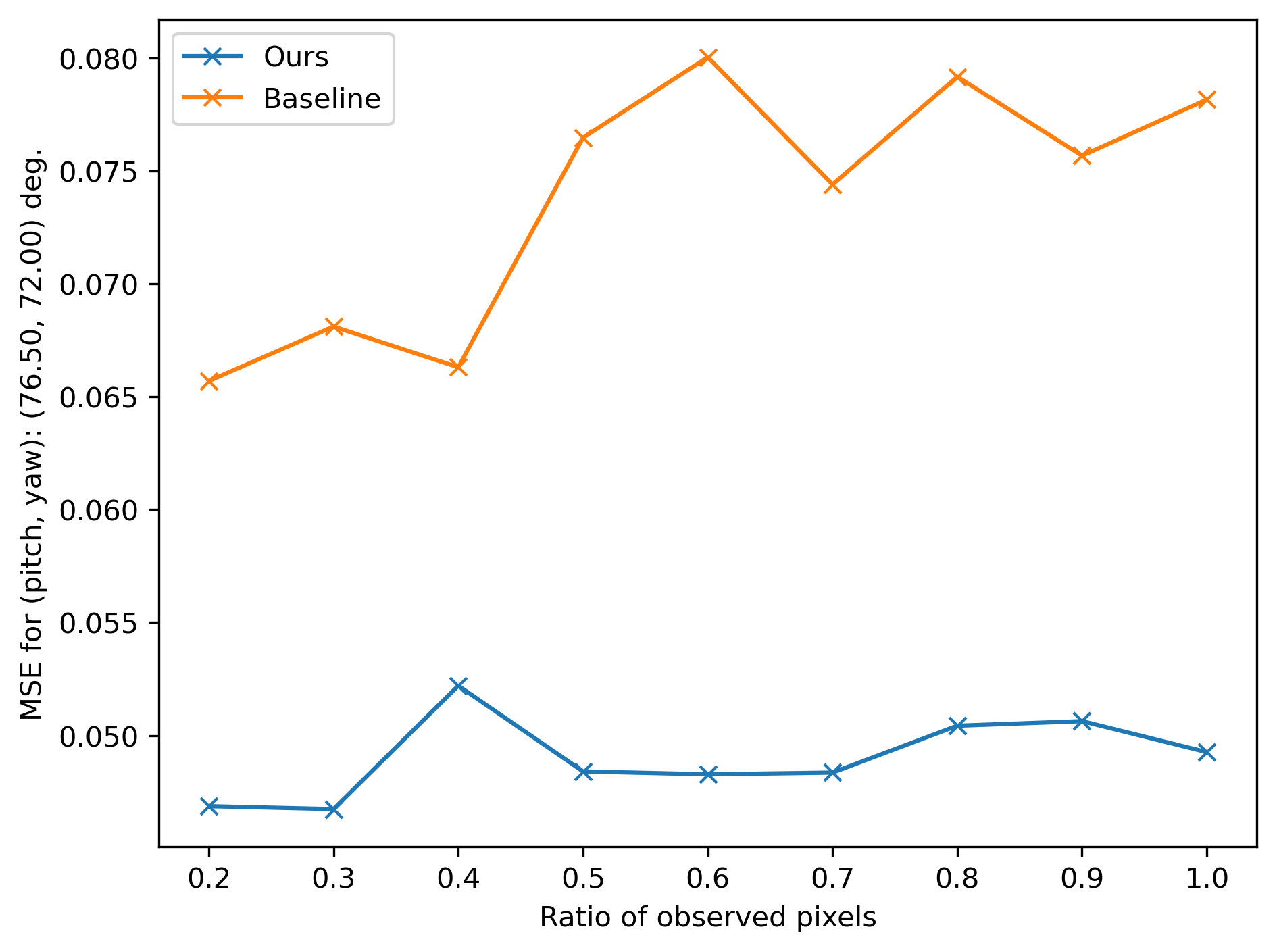}
    \end{tabular}
    \caption{Inpainting plots for different views, as we change the number of observed pixels in the frontal view. Our method consistently outperforms the baseline. As the number of observed pixels increases, the baseline method overfits more and more to the measurements (frontal view) and performs worse in novel views, indicating that the reconstructed geometry gets worse.}
    \label{fig:inpainting_plot}
\end{figure}

\section{Limitations}
The success of our method relies on the quality of the collected set of voxel grids. If the set is not diverse or if it contains non-realistic 3-D structures, then our method will fail for some instances. 
Also, since all the geometries are in the range of the GAN, any dataset biases will be reflected in our reconstructions. We note that our method might only introduce biases in the 3-D structure of a face since we are not regularizing for color. We plan to release our dataset of reference geometries for further inspection and we refer the readers to~\cite{jalal2021fairness, pulse, ilo} for a discussion on how GANs and different inversion algorithms can amplify dataset biases.

Another concern is that the relaxation of the optimization problem allows for solutions that are blendings of the 3-D structures of the collected set. A blending of two realistic facial geometries might have artifacts. We account for this by annealing the temperature, effectively encouraging the optimization to converge to a single 3-D structure. We do not have any principled way of annealing the temperature and we leave this as a future direction.

Finally, our method regularizes only for a realistic 3-D structure, but does not add any regularization on the colors. Non-smoothness in the 3-D color signal might give undesired transitions between nearby views. We do not observe any such behavior with the pi-GAN generator, but it could happen with other models from the NerfGAN family.

\section{Conclusions}
In this paper, we propose a novel method for solving inverse problems for 3-D neural radiance fields given a single 2-D view. The proposed framework naturally generalizes even if a partial or corrupted 2-D view is available and extends existing work on unsupervised methods for inverse problems. 
The central innovation is regularizing the neural radiance field using reference geometries and we expect that better references will improve our method. We expect our method to be applicable to generative neural radiance field methods and improve in performance as more powerful pre-trained NerfGANs become available.

\title{Supplementary Material \\ Solving Inverse Problems with NerfGANs}
\date{}
\author{Giannis Daras \\ {\tt\small giannisdaras@google.com} \and
Wen-Sheng Chu \\ {\tt\small wschu@google.com} \and
Abhishek Kumar \\ {\tt\small abhishk@google.com} \and
Dmitry Lagun \\ {\tt\small dlagun@google.com} \and
Alexandros G. Dimakis \\ {\tt\small dimakis@austin.utexas.edu}
}

\maketitle


\section{Additional Experiments}
This section provides additional experiments that could not fit in the paper due to the page limit. We first demonstrate that our method can be used for other objects, beyond human faces. The next step is to show how we can improve the quality of the renderings of our method by leveraging a 2-D powerful generator. We then show that our method is effective in solving general inverse problems, by obtaining realistic 3-D reconstructions given various types of image corruptions, such as compressed sensing, downsampling (super-resolution) and box inpainting. Finally, we present ablation studies that justify the design choices for our method.

\subsection{Results of our method for cats}
As discussed in the main paper, our method can work out-of-the-box for other objects, other than human faces. In this section, we provide evidence that supports this argument. Specifically, we use the pi-GAN generator pre-trained on the cats and we show that we render realistic novel views given a single image. 

One question is how we will obtain the reference geometries for the cats. We could follow the same procedure as the one mentioned in the paper, i.e. use CLIP~\cite{clip} to filter out bad geometries (of cats). However, by observing the generated geometries for random latents, we see that the cats pi-GAN generator very rarely outputs unrealistic 3-D structures in the unconditional generation task. Hence, we can sidestep the CLIP procedure and use random references instead. We use $32$ reference geometries for cats that correspond to latents sampled from the Gaussian distribution.

\begin{figure*}[!htp]
    \centering
    \begin{subfigure}[t]{0.2\textwidth}
            \begin{center}
                \includegraphics[width=\linewidth]{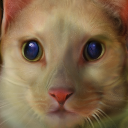}
                \caption{Reference}
            \end{center}
    \end{subfigure} \hspace{2mm} \vrule \hspace{2mm}
    \begin{subfigure}[t]{0.2\textwidth}
            \begin{center}
                \includegraphics[width=\linewidth]{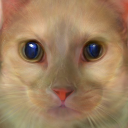}
                \caption{Novel view 1}
            \end{center}
    \end{subfigure}
    \begin{subfigure}[t]{0.2\textwidth}
            \begin{center}
                \includegraphics[width=\linewidth]{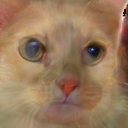}
                \caption{Novel view 2}
            \end{center}
    \end{subfigure}
    \begin{subfigure}[t]{0.2\textwidth}
            \begin{center}
                \includegraphics[width=\linewidth]{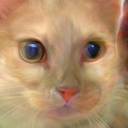}
                \caption{Novel view 3}
            \end{center}
    \end{subfigure}
    \begin{subfigure}[t]{\textwidth}
            \begin{center}
                \includegraphics[width=.5\linewidth]{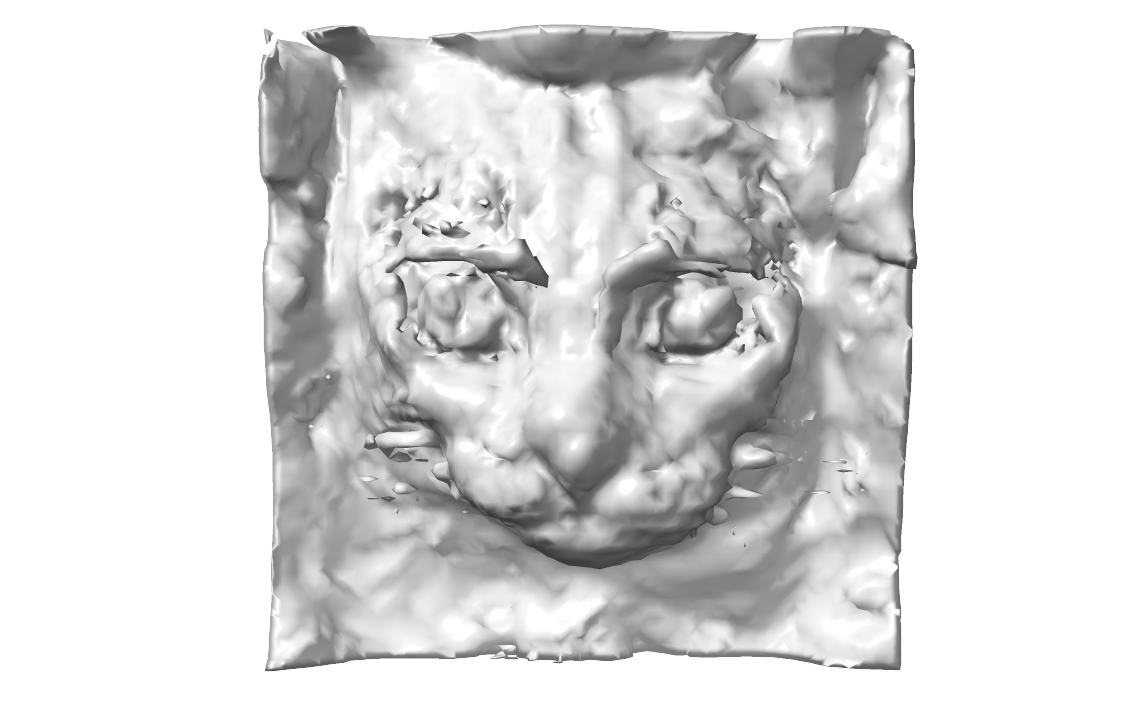}
                \caption{Reconstructed geometry}
            \end{center}
    \end{subfigure}
    \caption{{\bf Our results on cats:} 
    We show results with a generator trained on cats. Top row: Novel views generated by our method. Second row: Reconstructed geometry.}
    \label{fig:cats}
\end{figure*}

The results are shown in Figure \ref{fig:cats}. As seen, our method gives realistic renderings of novel views from a smooth 3-D geometry.

\subsection{Improving the reconstruction quality with Intermediate Layer Optimization}
In the main paper, we showed that our method removes 3-D obfuscations and hence gives more realistic novel views compared to the method described in pi-GAN~\cite{chan2021pi}. However, there are three issues that impact the quality of the novel views, even for our method:
\begin{enumerate}
    \item The pi-GAN generator is trained with low-resolution images. As a result, it cannot capture the fine details of real images.
    \item The pi-GAN generator (for faces) is trained to model only small camera movements. This is due to the fact that the CelebA~\cite{celeba} dataset has mostly frontal view images. Hence, when we move the camera in positions outside of the training distribution, the image quality deteriorates.
    \item The pi-GAN generator is not as powerful as state-of-the-art 2-D generators like StyleGAN~\cite{stylegan, stylegan2, stylegan3} and hence certain facial attributes cannot be modelled.
\end{enumerate}
These issues affect the image quality of our reconstructions. We can mitigate these issues using a 2-stage approach. At the very first stage, we use our method (described in the main paper) to obtain a coarse reconstruction and a smooth 3-D geometry. We use the radiance field from the solution of our optimization problem to render novel views. These views are expected to have blurriness and certain artifacts because of the pi-GAN issues described above. Then, we can use a powerful 2-D generator, \ie, StyleGAN, as a prior to correct these artifacts separately for every novel view.

One problem with projecting back to the range of StyleGAN is that the generator is not expressive enough to model all the images.
Hence we might get a photorealistic person that is not looking very close to our input (frontal view). Intermediate Layer Optimization (ILO)~\cite{ilo} solves this problem by optimizing in the intermediate layers of StyleGAN to increase the expressive power of the generator. 

\begin{figure*}[!htp]
    \centering
    \begin{subfigure}[t]{0.3\textwidth}
            \begin{center}
                \includegraphics[width=\linewidth]{images/biden.png}
                \caption{Reference}
            \end{center}
    \end{subfigure} \hspace{2mm} \vrule \hspace{2mm}
    \begin{subfigure}[t]{0.3\textwidth}
            \begin{center}
                \includegraphics[width=\linewidth]{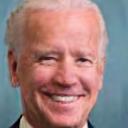}
                \caption{Novel view obtained with our method}
            \end{center}
    \end{subfigure} 
    \begin{subfigure}[t]{0.3\textwidth}
            \begin{center}
                \includegraphics[width=\linewidth]{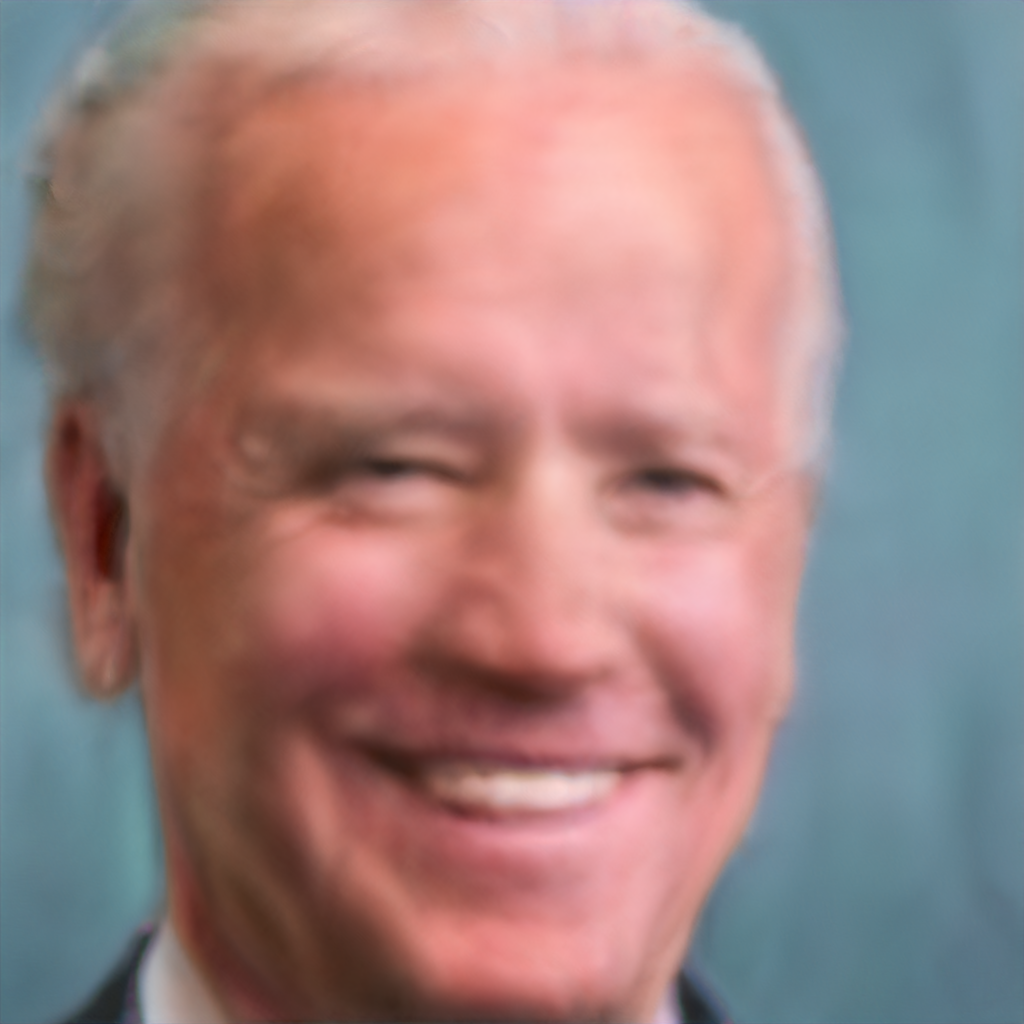}
                \caption{Enhanced with ILO~\cite{ilo}}
            \end{center}
    \end{subfigure} 
    \caption{{\bf Improved quality with ILO:}
    Novel view enhancement using StyleGAN~\cite{stylegan, stylegan2,stylegan3} and Intermediate Layer Optimization (ILO)~\cite{ilo}. We have a two-stages approach. We first use our method to generate consistent novel views and then we use ILO to project close to the StyleGAN range and enhance the image. Unnatural characteristics in the ears and the face have been removed because of the StyleGAN face prior.}
    \label{fig:ilo}
\end{figure*}

Figure \ref{fig:ilo} shows a novel view enhancement with ILO. Unnatural characteristics in the ears and the face have been removed because of the StyleGAN face prior. It is important to clarify that using ILO alone we could not generate novel views, since we do not have any control of the pose. Our method is essential as a first step in this process to reconstruct a smooth geometry and render consistent views that can later be enhanced with ILO or some other 2-D image restoration technology.

To accelerate the process, we can first solve for one novel view and then use this StyleGAN latent as a warm start for the optimization problem for a different view. In general, ILO~\cite{ilo} requires 1-2 mins per view on a single GPU.

We note that using ILO to enhance images generated by our method is only a trick and not the central innovation of our method. For this reason, we choose to not include it in the main paper. However, practitioners working with our method might find it useful to obtain better quality reconstructions, which is why we include it here.

An interesting future direction that is relevant to this experiment is learning a forward operator that maps from the actual novel views to the reconstructed novel views. Essentially, it would be useful to know in which way the images that we want to fix with ILO are corrupted and then adjust the ILO algorithm to better account for this corruption. We leave this as a future direction.

\subsection{Compressed Sensing}
We start our additional experiments by looking at the problem of Compressed Sensing. The goal here is to reconstruct a signal $x\in \R^n$ (frontal view) by observing some linear measurements $Ax$, where $A\in \R^{m\times n}$ is a gaussian i.i.d matrix. Since we are working with a NerfGAN, we solve the optimization problem of \ref{eq:gensol}.

\begin{figure}[!htp]
    \centering
    \includegraphics[width=\linewidth]{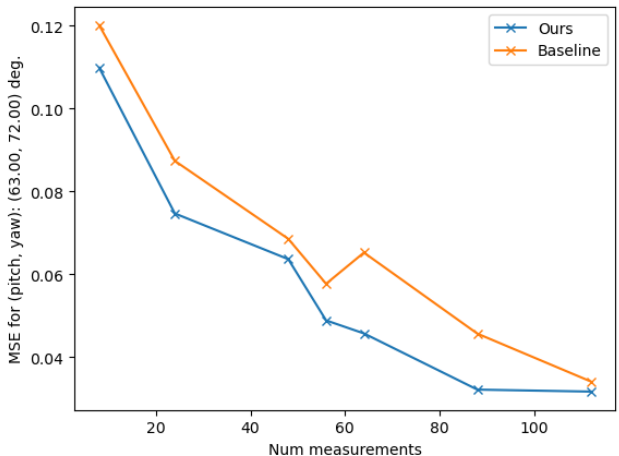}
    \caption{Quantitative results for an arbitrarily picked novel view for the task of \textbf{Compressed Sensing}. Our method consistently outperforms the baseline for all measurement settings. As the number of measurements increases, the error goes down as expected.}
    \label{fig:compressed_sensing_plot}
\end{figure}

Quantitative results for an arbitrarily picked novel view are shown in Figure \ref{fig:compressed_sensing_plot}. The Figure shows how the MSE error drops for both our method and the baseline as the number of measurements increase, as expected. Our method consistently outperforms the baseline for all measurements settings, which strengthens our argument that the 3D prior is useful for general inverse problems (\eg, super resolution, compressed sensing), and not just inversion and inpainting as shown in the main paper.

\begin{figure*}[!htp]
\centering
    \begin{subfigure}[t]{0.15\textwidth}
            \begin{center}
                \includegraphics[width=\linewidth]{images/gold/ones/view_1.0_1.0.jpg}
                \caption{Reference}
            \end{center}
    \end{subfigure} \hspace{2mm} \vrule \hspace{2mm}
    \begin{subfigure}[t]{0.15\textwidth}
            \begin{center}
                \includegraphics[width=\linewidth]{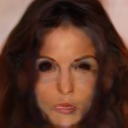}
                \caption{$m=16^2$}
            \end{center}
    \end{subfigure} 
    \begin{subfigure}[t]{0.15\textwidth}
            \begin{center}
                \includegraphics[width=\linewidth]{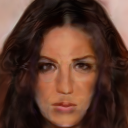}
                \caption{$m=24^2$}
            \end{center}
    \end{subfigure}  
    \begin{subfigure}[t]{0.15\textwidth}
            \begin{center}
                \includegraphics[width=\linewidth]{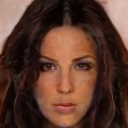}
                \caption{$m=32^2$}
            \end{center}
    \end{subfigure}  
    \begin{subfigure}[t]{0.15\textwidth}
            \begin{center}
                \includegraphics[width=\linewidth]{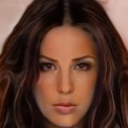}
                \caption{$m=128^2$}
            \end{center}
    \end{subfigure}
    \caption{Effect of number of measurements in reconstructed images for the task of compressed sensing. We want to reconstruct the reference image $x\in \R^{n}$, by observing $y=Ax$, where $A$ is a linear Gaussian matrix $\in \R^{m\times n}$. For the reference image, we have $n=128^2 \times 3$. As shown, increasing the number of measurements $m$, leads to better reconstruction of the reference image.}
    \label{fig:compressed_sensing_images}
\end{figure*}

To get a sense of how the number of measurements impacts the quality of the reconstructed image, we show examples of reconstructions for different measurement settings with our method. The images are shown in Figure \ref{fig:compressed_sensing_images}. It is evident that as the number of measurements $m$ increases, our reconstruction of the ground truth becomes more accurate, in agreement with the MSE plot of Figure \ref{fig:compressed_sensing_plot}.

\subsection{Super-resolution}
\begin{figure}[!htp]
    \centering
    \includegraphics[width=\linewidth]{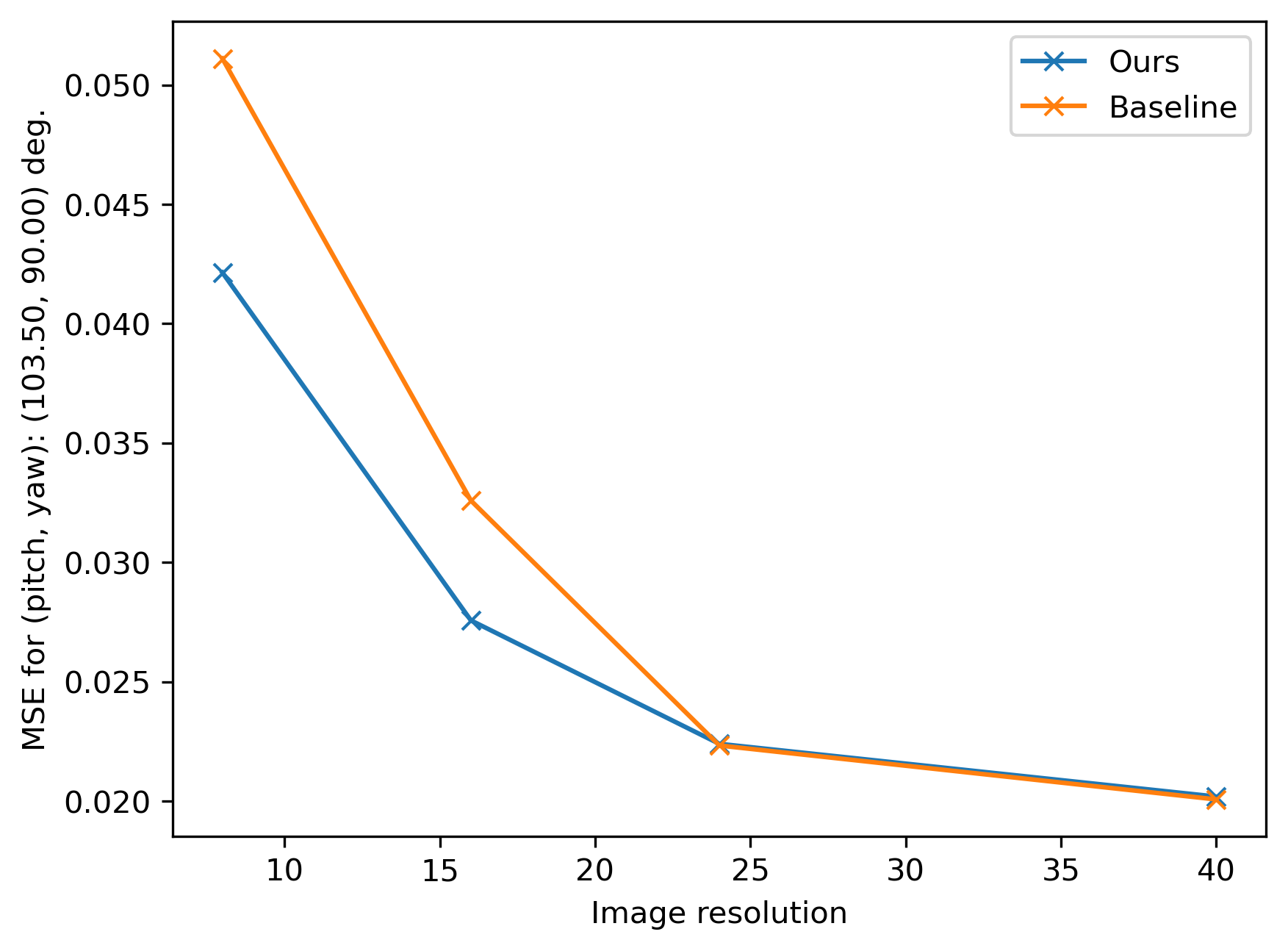}
    \caption{Quantitative results for an arbitrarily picked novel view for the task of \textbf{Super Resolution}. Our method outperforms the baseline in the low-measurements and matches its performance everywhere else. As the number of measurements increases, the error goes down as expected.}
    \label{fig:super_res_plot}
\end{figure}

The next task we consider is the super-resolution task where the goal is to infer a higher-resolution version of a given image (and render novel views with it) by observing a very pixelated, low-resolution version of it.
Quantitative results for an arbitrarily picked novel view for this task are shown in Figure \ref{fig:super_res_plot}. Our method outperforms the baseline in the low-measurements and matches its performance everywhere else. As the number of measurements increases, the error goes down as expected (similar to what happened for the Compressed Sensing case).

\begin{figure*}
    \centering
    \begin{subfigure}[t]{0.15\textwidth}
            \begin{center}
                \includegraphics[width=\linewidth]{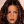}
                \caption{Input image}
            \end{center}
    \end{subfigure} \hspace{2mm} \vrule \hspace{2mm}
    \begin{subfigure}[t]{0.15\textwidth}
            \begin{center}
                \includegraphics[width=\linewidth]{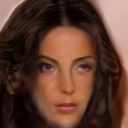}
                \caption{Novel view rendered with our method.}
            \end{center}
    \end{subfigure} 
    \begin{subfigure}[t]{0.15\textwidth}
            \begin{center}
                \includegraphics[width=\linewidth]{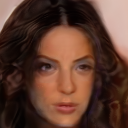}
                \caption{Baseline novel view.}
            \end{center}
    \end{subfigure} 
    \caption{Super-resolution visual results. Given a low-resolution image (first column, resolution $16\times 16$) we show novel views rendered by our method and the baseline at resolution $128\times 128$. }
    \label{fig:super_res}
\end{figure*}

Visual results for the task of super-resolution our shown in Figure \ref{fig:super_res}. Given a low-resolution image (first column, resolution $16\times 16$) we show novel views rendered by our method and the baseline at resolution $128\times 128$. Both methods seem to be performing relatively well on this task, but as the MSE scores of Figure \ref{fig:super_res_plot}, our method has better performance in the very low-measurements regime.

\subsection{Inpainting}
\begin{figure*}
    \centering
    \begin{subfigure}[t]{0.2\textwidth}
            \begin{center}
                \includegraphics[width=\linewidth]{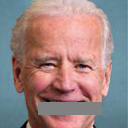}
                \caption{Input image}
            \end{center}
    \end{subfigure} \hspace{2mm} \vrule \hspace{2mm}
    \begin{subfigure}[t]{0.2\textwidth}
            \begin{center}
                \includegraphics[width=\linewidth]{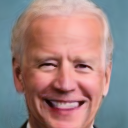}
                \caption{Novel view}
            \end{center}
    \end{subfigure} 
    \begin{subfigure}[t]{0.2\textwidth}
            \begin{center}
                \includegraphics[width=\linewidth]{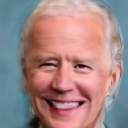}
                \caption{Novel view}
            \end{center}
    \end{subfigure} 
    \begin{subfigure}[t]{0.2\textwidth}
            \begin{center}
                \includegraphics[width=\linewidth]{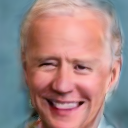}
                \caption{Novel view}
            \end{center}
    \end{subfigure} 
    \caption{Inpainting of an image box. Our method is capable of filling the missing region in a plausible way.}
    \label{fig:box_inp}
\end{figure*}

In the paper, we presented results for the task of randomized $2$-D inpainting. Figure \ref{fig:box_inp} illustrates one example of box inpainting. As shown, our method is capable of filling the missing region in a plausible way.

\subsection{Ablation: number of reference geometries}
We run an ablation study to examine the role of the number of reference geometries in the quality of the reconstructions. Intuitively, we expect that as we increase the number of reference geometries, the trend should be that the MSE should go down for novel views. This follows from the fact that our annealing scheme will force the optimization to converge to a single reference geometry eventually, so as the set of reference geometries becomes bigger, we expeect that we can find a geometry that matches our measurements \textit{and} generalizes well to other views. 

\begin{figure}
    \centering
    \includegraphics[width=\linewidth]{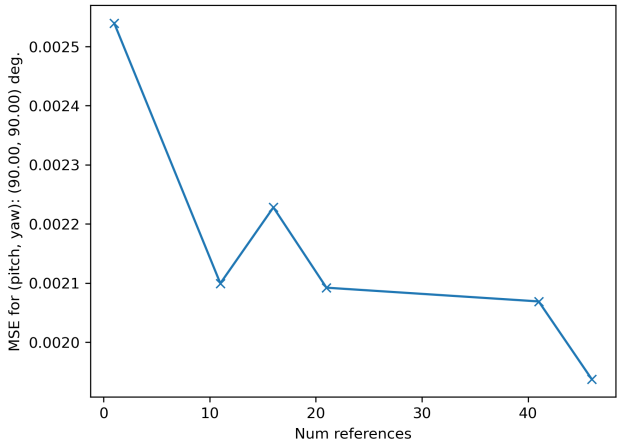}
    \caption{Ablation on how the number of reference grids affects the MSE error. The observed trend is that as the number of reference geometries goes up, the MSE error goes down, even for the frontal view. This agrees with our intuition that a more diverse set of reference geometries gives more flexibility to the optimization procedure to discover a geometry that matches well the measurements. For this experiment, we used Tempetature Annealing, as described in the paper.}
    \label{fig:voxel_ablation}
\end{figure}

Our intuition is confirmed by Figure \ref{fig:voxel_ablation}. The Figure shows that increasing the number of reference voxel grids leads to lower MSE, even in the frontal view.  A more diverse set of reference geometries gives more flexibility to the optimization procedure to discover a geometry that matches well the measurements

The trade-off here is that as the number of reference geometries increases, the computational cost of running the method increases as well (since we have to compute MSE between the current voxel grid and all the reference voxel grids). There are ways to mitigate the computational issue (i.e. completely removing some geometries as the inverse temperature goes up), but we will leave that as a future work.

\subsection{Ablation: comparison with piGAN's regularization}
pi-GAN~\cite{chan2021pi} optimizes over the frequencies and the phase-shifts over the SIREN~\cite{siren} network.  During training, the frequencies and the phase shifts for all SIREN layers are the same. When solving inverse problems, the authors of~\cite{chan2021pi} propose to let them diverge, in an attempt to make the generator more powerful. This is very similar in spirit in how StyleGAN~\cite{stylegan2} optimizes over the styles for all layers.

The main difference is that StyleGAN enforces a geodesic loss while pi-GAN penalizes frequencies and phase shifts for getting away from their average values. Figure 2 of the main paper shows that this regularization significantly limits the power of the generator (images are taken directly from the pi-GAN paper). In this section, we explore this observation a little bit more. Specifically, we run pi-GAN's proposed regularization, geodesic regularization and our method (3-D loss + geodesic) for real images from CelebA~\cite{celeba} and we report the average MSE scores in Table \ref{tab:celeba_scores}. As shown, Geodesic Regularization and the pi-GAN's regularization perform on par and our method consistently outperforms both of them by a large margin.

\begin{table}[!htp]
    \centering
    \begin{tabular}{c|c|c}
        & Frontal view MSE & Frontal view LPIPS \\ \hline
         pi-GAN &  0.0070 & 89.95 \\
         Geodesic & 0.0068 & 94.01 \\
         Ours & \textbf{0.0037} & \textbf{36.17} \\
    \end{tabular}
    \caption{Comparison of different regularization schemes on real images from CelebA~\cite{celeba}.}
    \label{tab:celeba_scores}
\end{table}

\section{Discussion}
\subsection{Things that did not work}
In this section, we share some negative results. Our goal is to make our efforts known to other researchers working in this field so they can avoid our methods, adjust them or even contradict our findings.

\paragraph{3-D Loss functions} We spent a lot of time deriving a 3-D loss function that penalizes unrealistic voxel grids. We ended up defining something as unrealistic if it is far from a (known to be good) set of reference geometries. However, initially, we tried removing the dependency on the 3-D geometries and just regularize the 3-D voxel grid by penalizing unexpected behaviors. 

Outside of the facial area, we expect the density values to be small since they correspond to empty space. Hence, a natural regularization would minimize the $l_2$ norm of the vectors with the densities of the non-facial surface. This method failed miserably. We attribute the failure to specific patterns that usually appear in the pi-GAN generated voxel grids. Specifically, the $l_2$ regularization encourages all the selected voxels to have small densities independently. pi-GAN does not have this structure; periodic patterns appear in the densities that most likely stem from the frequencies we feed to the SIREN~\cite{siren} network. 

Another property that we tried to enforce is smoothness of the 3-D signal, \ie, Lipschitzness of the 3-D gradients. Our motivation was to reduce sudden changes in the facial structure that appeared when moving the camera even a little bit from the frontal view. Optimizing for this objective led to a different type of unrealistic geometries, where whole areas were smoothed out and a few spikes in the frontal view captured the information needed for a decent frontal view rendering.

Finally, we tried removing the dependency on the facial masks (obtained using the Marching Cubes algorithm). Our goal was to penalize unexpected behaviors everywhere, not just outside of the face. This method also failed; the expressivity of the generator reduced dramatically when we tried to match all the reference voxels. In retrospect, we could have expected that since this is esentially trying to match a random 3-D face from a dataset of 3-D faces.


\paragraph{Optimization} Any method that regularizes inverse problems shows a trade-off between matching the measurements (in our case, the frontal view) and respecting the prior (in our case, maintaining a realistic 3-D structure). We explored different ways to balance this trade-off by applying optimization tricks. Specifically, we tried to do Alternating Gradient Descent~\cite{wright2015coordinate}, i.e. do one optimization step to minimize the measurements and one to respect the prior. We experimented with unbalanced Alternating Gradient Descent (e.g. taking more steps to fit the measurements) and we also tried to first fit the measurements completely and then adjust to respect the prior. All these attempts improved only marginally the results and hence they are not used in the paper to avoid the complication of tuning them for all our experiments. Finally, we tried different learning rates schedulings, but we observed no big improvement to the results over the vanilla Adam~\cite{kingma2017adam} optimizer with weight decay.

\subsection{CLIP}
In the main paper, we explained that we use CLIP~\cite{clip} as a way to automatically filter out bad geometries and consequently form a set of reference geometries. In this section, we provide more details about how this is done in practice and we include examples of bad geometries and their corresponding 2-D views, as detected by CLIP.

The main idea is that we expect consistent geometries to generate views that are not very different in some semantically meaningful space, such as the space of CLIP embeddings. In all our experiments, we render 9 views with CLIP, with $\theta \in \{76.5^\circ, 90^\circ, 103.5^\circ\}$ and $\phi \in \{81^\circ, 90^\circ, 99^\circ \}$. We found this value by experimenting and manually inspecting the CLIP classifications of good and bad geometries for $10$ validation images.

\begin{figure}[!htp]
    \centering
    \begin{subfigure}[t]{0.12\textwidth}
            \begin{center}
                \includegraphics[width=\linewidth]{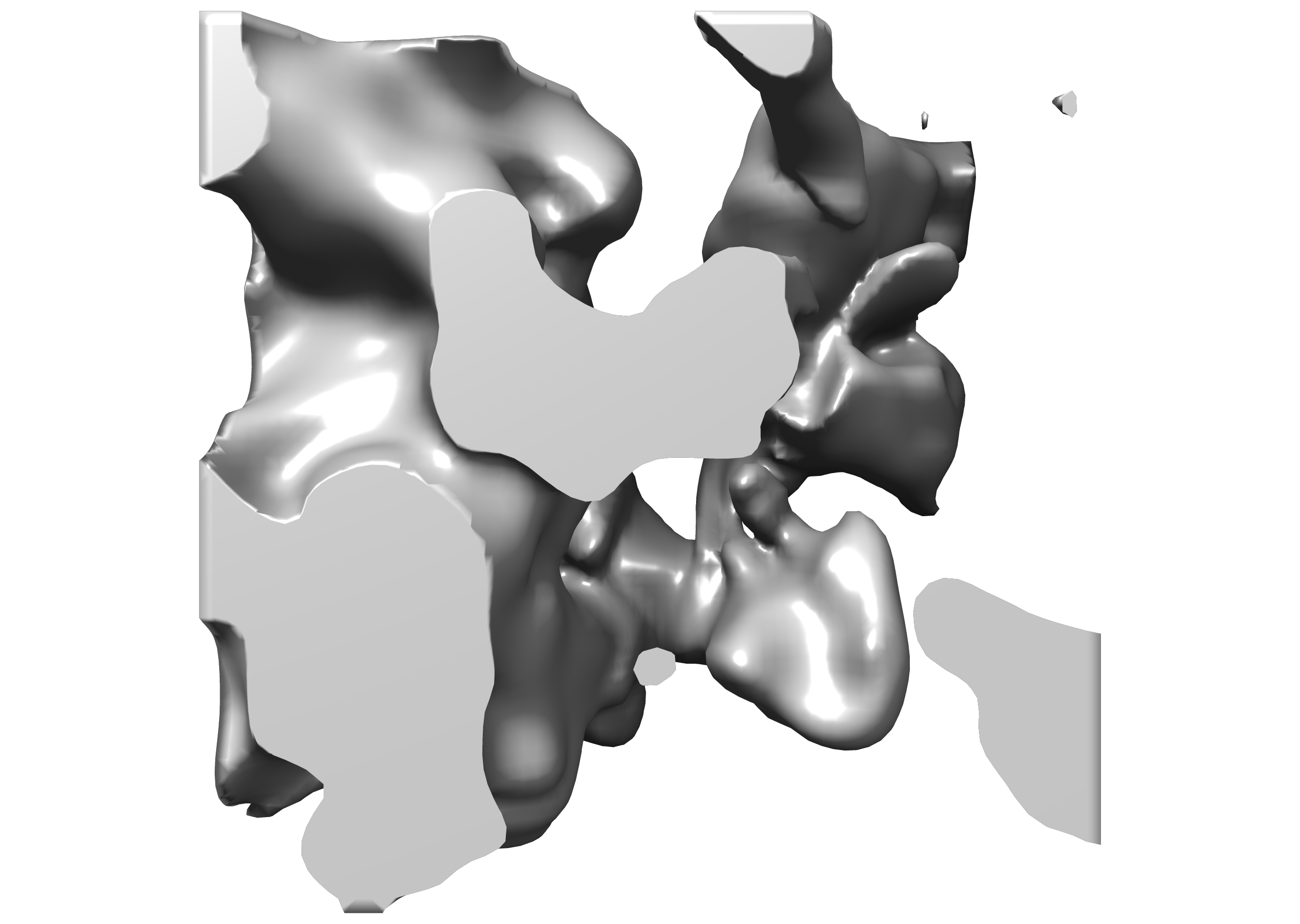}
            \end{center}
    \end{subfigure}
    \begin{subfigure}[t]{0.12\textwidth}
            \begin{center}
                \includegraphics[width=\linewidth]{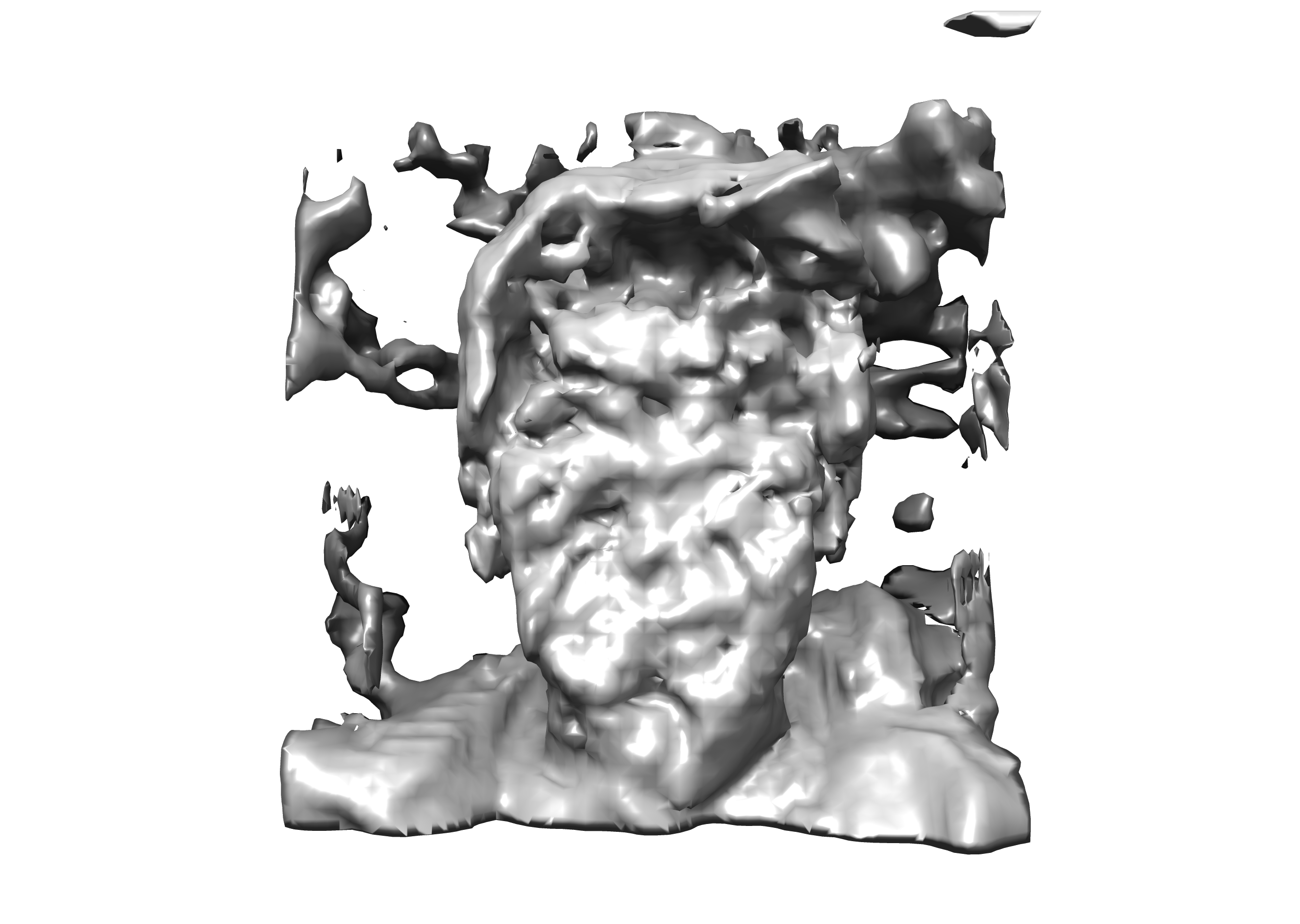}
            \end{center}
    \end{subfigure}
    \begin{subfigure}[t]{0.12\textwidth}
            \begin{center}
                \includegraphics[width=\linewidth]{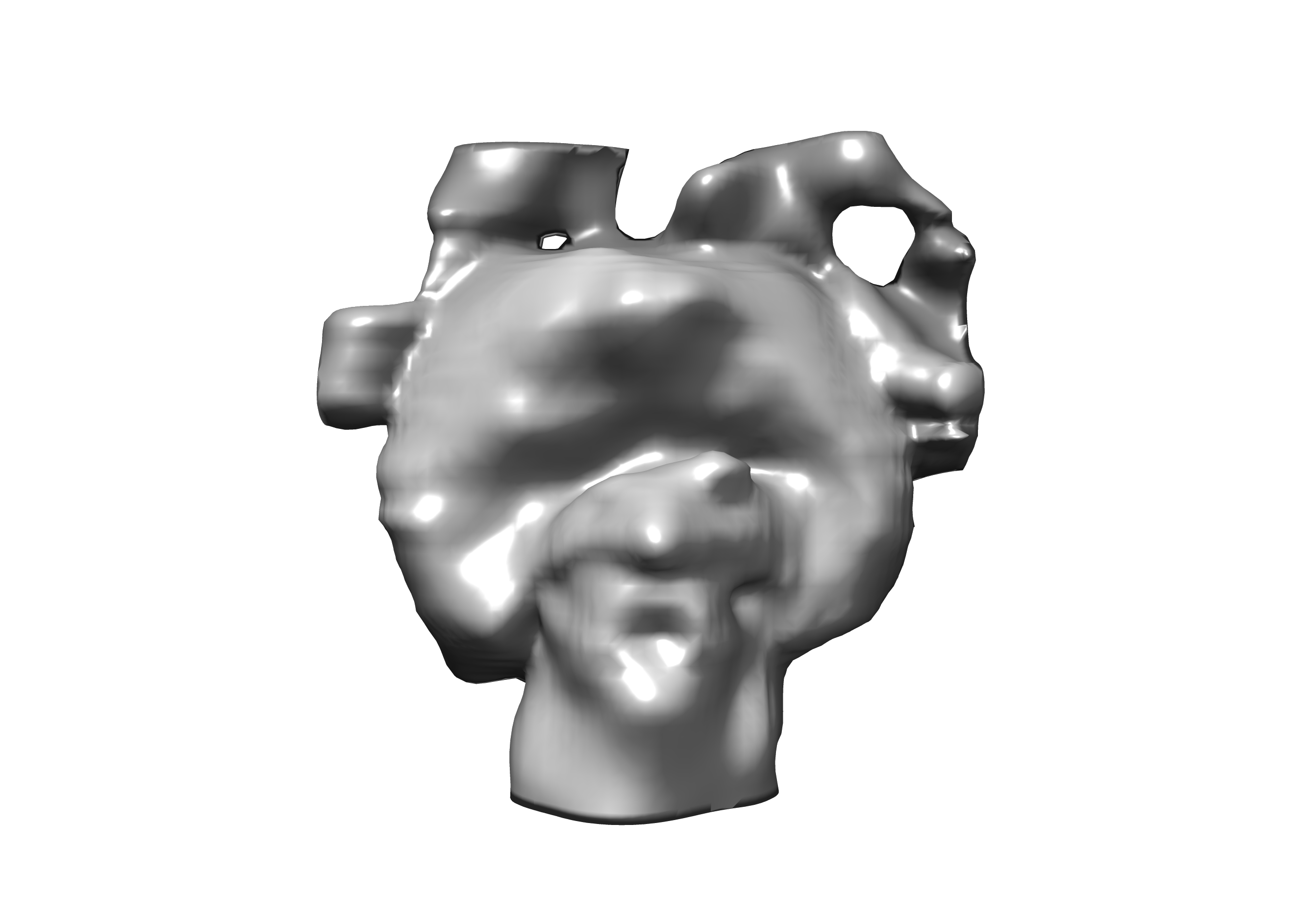}
            \end{center}
    \end{subfigure}
    \begin{subfigure}[t]{0.12\textwidth}
            \begin{center}
                \includegraphics[width=\linewidth]{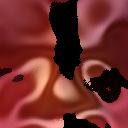}
            \end{center}
    \end{subfigure}
    \begin{subfigure}[t]{0.12\textwidth}
            \begin{center}
                \includegraphics[width=\linewidth]{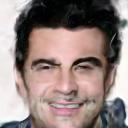}
            \end{center}
    \end{subfigure}
    \begin{subfigure}[t]{0.12\textwidth}
            \begin{center}
                \includegraphics[width=\linewidth]{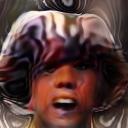}
            \end{center}
    \end{subfigure}
    \caption{Automatically classified bad geometries and their corresponding 2-D views. For the classification, we used the free-text classifier, CLIP~\cite{clip}. Surprisingly, the collected geometries are produced by latents sampled from the Gaussian distribution. This indicates that pi-GAN has failure modes that we need to avoid when solving inverse problems.}
    \label{fig:clip}
\end{figure}

Examples of identified bad geometries and their corresponding 2-D images are shown in Figure \ref{fig:clip}.
Surprisingly, the collected geometries are pro-duced by latents sampled from the Gaussian distribution. This in-dicates that pi-GAN has failure modes that we need to avoid whensolving inverse problems

\section{Experimental Details}
\subsection{Hyperparameters}
In all our experiments, we tried to follow as closely as possible the hyperparameters reported in the pi-GAN paper~\cite{chan2021pi}. For the baseline runs, we started with the hyperparameters reported in the paper and we reported the best result among this run and our tuning of the baseline's hyperparameters. 

We use the Adam~\cite{kingma2017adam} optimizer with initial learning rate at $0.01$ and a half step decay every $200$ steps, as recommended. As in the pi-GAN paper, for all inverse problems, we optimize for the frequencies and the phase-shifts (and not the latent $z$ itself). During training, the model had the same frequencies and phase-shifts across all layers for a single image. Inspired by the StyleGAN~\cite{stylegan, stylegan2, stylegan3} papers we allow these frequencies and phase-shifts to diverge. The pi-GAN paper is using MSE from their average values as regularization that forces them to stay close. In our paper, we use Geodesic Loss, as used in~\cite{stylegan, pulse, ilo}.

Additionally to the MSE distance, we also use a Perceptual Distance, LPIPS~\cite{lpips}, for the inversion problem, as in~\cite{ilo}. For a fair comparison with the baseline, we have deactivated LPIPS for all the plots included in the paper. However, for best visual results, we recommend adding the Perceptual Loss in addition to MSE.

Below, we provide a list of hyperparameters that one needs to tune for best results with method and a proposed set of sensible values.

\begin{itemize}
    \item Temperature Annealing Values: $[100., 150., 200., 250., 300., 350., 400., 500., 550.]$
    \item Temperature Annealing Steps: $[100, 200, 300, 400, 500, 600, 700]$
    \item Number of reference voxel grids: $\{16, 32, 64, 128\}$
    \item Voxel grid resolution: $\{32^3, 64^4\}$
    \item Learning rate: \{1e-2, 5e-3, 1e-3\}
    \item LPIPS coefficient: $\{1e-2, 5e-3\}$
    \item MSE coefficient: $\{1, 5, 10\}$
    \item Prior coefficient: $\{1, 1e-1, 1e-2\}$
    \item Geodesic coefficient: $\{5e-1, 1e-1\}$
\end{itemize}

\subsection{Plots}
For all the plots, we are running our method and the baseline for $100$ images in the range of the GAN and we average across all runs. For the inverse problems where corruption has taken place (e.g. inpainting, super-resolution) we report MSE to the ground-truth signal that is never observed during the optimization. For all these plots, we are able to report MSE for novel views because the images we test on are in the range of the GAN, i.e. we first sampled a latent $z$ and then produced them. For real images, we can only compare the visual results between our method and the baseline as we did in the paper.

{\small
\bibliographystyle{ieee_fullname} 
\bibliography{egbib}}

\begin{thebibliography}{10}\itemsep=-1pt

\bibitem{bora2017compressed}
Ashish Bora, Ajil Jalal, Eric Price, and Alexandros~G Dimakis.
\newblock Compressed sensing using generative models.
\newblock In {\em International Conference on Machine Learning}, pages
  537--546. PMLR, 2017.

\bibitem{biggan}
Andrew Brock, Jeff Donahue, and Karen Simonyan.
\newblock Large scale gan training for high fidelity natural image synthesis,
  2018.

\bibitem{chan2021pi}
Eric~R Chan, Marco Monteiro, Petr Kellnhofer, Jiajun Wu, and Gordon Wetzstein.
\newblock pi-gan: Periodic implicit generative adversarial networks for
  3d-aware image synthesis.
\newblock In {\em Proceedings of the IEEE/CVF Conference on Computer Vision and
  Pattern Recognition}, pages 5799--5809, 2021.

\bibitem{ilo}
Giannis Daras, Joseph Dean, Ajil Jalal, and Alexandros~G Dimakis.
\newblock Intermediate layer optimization for inverse problems using deep
  generative models.
\newblock {\em arXiv preprint arXiv:2102.07364}, 2021.

\bibitem{daras2020smyrf}
Giannis Daras, Nikita Kitaev, Augustus Odena, and Alexandros~G Dimakis.
\newblock Smyrf: Efficient attention using asymmetric clustering.
\newblock {\em arXiv preprint arXiv:2010.05315}, 2020.

\bibitem{ylg}
Giannis Daras, Augustus Odena, Han Zhang, and Alexandros~G Dimakis.
\newblock Your local gan: Designing two dimensional local attention mechanisms
  for generative models.
\newblock In {\em Proceedings of the IEEE/CVF Conference on Computer Vision and
  Pattern Recognition}, pages 14531--14539, 2020.

\bibitem{Du20arxiv_nerflow}
Yilun Du, Yinan Zhang, Hong-Xing Yu, Joshua~B. Tenenbaum, and Jiajun Wu.
\newblock Neural radiance flow for {4D} view synthesis and video processing.
\newblock {\em arXiv preprint arXiv:2012.09790}, 2020.

\bibitem{3dmm}
Bernhard Egger, William A.~P. Smith, Ayush Tewari, Stefanie Wuhrer, Michael
  Zollhoefer, Thabo Beeler, Florian Bernard, Timo Bolkart, Adam Kortylewski,
  Sami Romdhani, and et al.
\newblock 3d morphable face models—past, present, and future.
\newblock {\em ACM Transactions on Graphics}, 39(5):1–38, Sep 2020.

\bibitem{esser2021taming}
Patrick Esser, Robin Rombach, and Bjorn Ommer.
\newblock Taming transformers for high-resolution image synthesis.
\newblock In {\em Proceedings of the IEEE/CVF Conference on Computer Vision and
  Pattern Recognition}, pages 12873--12883, 2021.

\bibitem{Garbin21arxiv_FastNeRF}
Stephan~J. Garbin, Marek Kowalski, Matthew Johnson, Jamie Shotton, and Julien
  Valentin.
\newblock Fastnerf: High-fidelity neural rendering at 200fps.
\newblock {\em https://arxiv.org/abs/2103.10380}, 2021.

\bibitem{goodfellow2014generative}
Ian Goodfellow, Jean Pouget-Abadie, Mehdi Mirza, Bing Xu, David Warde-Farley,
  Sherjil Ozair, Aaron Courville, and Yoshua Bengio.
\newblock Generative adversarial nets.
\newblock {\em Advances in neural information processing systems}, 27, 2014.

\bibitem{gu2021stylenerf}
Jiatao Gu, Lingjie Liu, Peng Wang, and Christian Theobalt.
\newblock Stylenerf: A style-based 3d-aware generator for high-resolution image
  synthesis, 2021.

\bibitem{hand2018phase}
Paul Hand, Oscar Leong, and Vladislav Voroninski.
\newblock Phase retrieval under a generative prior.
\newblock {\em arXiv preprint arXiv:1807.04261}, 2018.

\bibitem{jalal2021mri}
Ajil Jalal, Marius Arvinte, Giannis Daras, Eric Price, Alexandros~G Dimakis,
  and Jonathan~I Tamir.
\newblock Robust compressed sensing mri with deep generative priors.
\newblock In {\em Advances in Neural Information Processing Systems (NeurIPS)},
  2021.

\bibitem{jalal2021fairness}
Ajil Jalal, Sushrut Karmalkar, Jessica Hoffmann, Alex Dimakis, and Eric Price.
\newblock Fairness for image generation with uncertain sensitive attributes.
\newblock In {\em International Conference on Machine Learning}, pages
  4721--4732. PMLR, 2021.

\bibitem{stylegan3}
Tero Karras, Miika Aittala, Samuli Laine, Erik Härkönen, Janne Hellsten,
  Jaakko Lehtinen, and Timo Aila.
\newblock Alias-free generative adversarial networks, 2021.

\bibitem{stylegan}
Tero Karras, Samuli Laine, and Timo Aila.
\newblock A style-based generator architecture for generative adversarial
  networks, 2018.

\bibitem{stylegan2}
Tero Karras, Samuli Laine, Miika Aittala, Janne Hellsten, Jaakko Lehtinen, and
  Timo Aila.
\newblock Analyzing and improving the image quality of stylegan.
\newblock In {\em Proceedings of the IEEE/CVF Conference on Computer Vision and
  Pattern Recognition}, pages 8110--8119, 2020.

\bibitem{kingma2017adam}
Diederik~P. Kingma and Jimmy Ba.
\newblock Adam: A method for stochastic optimization, 2017.

\bibitem{kumar2021constrained}
Abhishek Kumar and Ehsan Amid.
\newblock Constrained instance and class reweighting for robust learning under
  label noise, 2021.

\bibitem{li2021neural}
Tianye Li, Mira Slavcheva, Michael Zollhoefer, Simon Green, Christoph Lassner,
  Changil Kim, Tanner Schmidt, Steven Lovegrove, Michael Goesele, and Zhaoyang
  Lv.
\newblock Neural 3d video synthesis, 2021.

\bibitem{Li20arxiv_nsff}
Zhengqi Li, Simon Niklaus, Noah Snavely, and Oliver Wang.
\newblock Neural scene flow fields for space-time view synthesis of dynamic
  scenes.
\newblock {\em https://arxiv.org/abs/2011.13084}, 2020.

\bibitem{Lindell20arxiv_AutoInt}
David Lindell, Julien Martel, and Gordon Wetzstein.
\newblock {AutoInt}: Automatic integration for fast neural volume rendering.
\newblock {\em https://arxiv.org/abs/2012.01714}, 2020.

\bibitem{liu2020neural}
Lingjie Liu, Jiatao Gu, Kyaw~Zaw Lin, Tat-Seng Chua, and Christian Theobalt.
\newblock Neural sparse voxel fields.
\newblock {\em arXiv preprint arXiv:2007.11571}, 2020.

\bibitem{celeba}
Ziwei Liu, Ping Luo, Xiaogang Wang, and Xiaoou Tang.
\newblock Deep learning face attributes in the wild.
\newblock In {\em Proceedings of International Conference on Computer Vision
  (ICCV)}, December 2015.

\bibitem{liu2018large}
Ziwei Liu, Ping Luo, Xiaogang Wang, and Xiaoou Tang.
\newblock Large-scale celebfaces attributes (celeba) dataset.
\newblock {\em Retrieved August}, 15(2018):11, 2018.

\bibitem{lorensen1987marching}
William~E Lorensen and Harvey~E Cline.
\newblock Marching cubes: A high resolution 3d surface construction algorithm.
\newblock {\em ACM siggraph computer graphics}, 21(4):163--169, 1987.

\bibitem{meng2021gnerf}
Quan Meng, Anpei Chen, Haimin Luo, Minye Wu, Hao Su, Lan Xu, Xuming He, and
  Jingyi Yu.
\newblock Gnerf: Gan-based neural radiance field without posed camera.
\newblock {\em arXiv preprint arXiv:2103.15606}, 2021.

\bibitem{pulse}
Sachit Menon, Alexandru Damian, Shijia Hu, Nikhil Ravi, and Cynthia Rudin.
\newblock Pulse: Self-supervised photo upsampling via latent space exploration
  of generative models.
\newblock {\em 2020 IEEE/CVF Conference on Computer Vision and Pattern
  Recognition (CVPR)}, Jun 2020.

\bibitem{nerf}
Ben Mildenhall, Pratul~P Srinivasan, Matthew Tancik, Jonathan~T Barron, Ravi
  Ramamoorthi, and Ren Ng.
\newblock Nerf: Representing scenes as neural radiance fields for view
  synthesis.
\newblock In {\em European conference on computer vision}, pages 405--421.
  Springer, 2020.

\bibitem{neff2021donerf}
Thomas Neff, Pascal Stadlbauer, Mathias Parger, Andreas Kurz, Joerg~H Mueller,
  Chakravarty R~Alla Chaitanya, Anton Kaplanyan, and Markus Steinberger.
\newblock Donerf: Towards real-time rendering of compact neural radiance fields
  using depth oracle networks.
\newblock {\em arXiv preprint arXiv:2103.03231}, 2021.

\bibitem{nguyenphuoc2018rendernet}
Thu Nguyen-Phuoc, Chuan Li, Stephen Balaban, and Yong-Liang Yang.
\newblock Rendernet: A deep convolutional network for differentiable rendering
  from 3d shapes, 2018.

\bibitem{nguyen2019hologan}
Thu Nguyen-Phuoc, Chuan Li, Lucas Theis, Christian Richardt, and Yong-Liang
  Yang.
\newblock Hologan: Unsupervised learning of 3d representations from natural
  images.
\newblock In {\em Proceedings of the IEEE/CVF International Conference on
  Computer Vision}, pages 7588--7597, 2019.

\bibitem{niemeyer2021giraffe}
Michael Niemeyer and Andreas Geiger.
\newblock Giraffe: Representing scenes as compositional generative neural
  feature fields.
\newblock In {\em Proceedings of the IEEE/CVF Conference on Computer Vision and
  Pattern Recognition}, pages 11453--11464, 2021.

\bibitem{ongie2020deep}
Gregory Ongie, Ajil Jalal, Christopher~A Metzler, Richard~G Baraniuk,
  Alexandros~G Dimakis, and Rebecca Willett.
\newblock Deep learning techniques for inverse problems in imaging.
\newblock {\em IEEE Journal on Selected Areas in Information Theory},
  1(1):39--56, 2020.

\bibitem{pan2021shadingguided}
Xingang Pan, Xudong Xu, Chen~Change Loy, Christian Theobalt, and Bo Dai.
\newblock A shading-guided generative implicit model for shape-accurate
  3d-aware image synthesis, 2021.

\bibitem{clip}
Alec Radford, Jong~Wook Kim, Chris Hallacy, Aditya Ramesh, Gabriel Goh,
  Sandhini Agarwal, Girish Sastry, Amanda Askell, Pamela Mishkin, Jack Clark,
  Gretchen Krueger, and Ilya Sutskever.
\newblock Learning transferable visual models from natural language
  supervision, 2021.

\bibitem{dalle}
Aditya Ramesh, Mikhail Pavlov, Gabriel Goh, Scott Gray, Chelsea Voss, Alec
  Radford, Mark Chen, and Ilya Sutskever.
\newblock Zero-shot text-to-image generation, 2021.

\bibitem{Rebain20arxiv_derf}
Daniel Rebain, Wei Jiang, Soroosh Yazdani, Ke Li, Kwang~Moo Yi, and Andrea
  Tagliasacchi.
\newblock {DeRF}: Decomposed radiance fields.
\newblock {\em https://arxiv.org/abs/2011.12490}, 2020.

\bibitem{reiser2021kilonerf}
Christian Reiser, Songyou Peng, Yiyi Liao, and Andreas Geiger.
\newblock Kilonerf: Speeding up neural radiance fields with thousands of tiny
  mlps.
\newblock {\em arXiv preprint arXiv:2103.13744}, 2021.

\bibitem{schwarz2021graf}
Katja Schwarz, Yiyi Liao, Michael Niemeyer, and Andreas Geiger.
\newblock Graf: Generative radiance fields for 3d-aware image synthesis, 2021.

\bibitem{siren}
Vincent Sitzmann, Julien N.~P. Martel, Alexander~W. Bergman, David~B. Lindell,
  and Gordon Wetzstein.
\newblock Implicit neural representations with periodic activation functions,
  2020.

\bibitem{wright2015coordinate}
Stephen~J Wright.
\newblock Coordinate descent algorithms.
\newblock {\em Mathematical Programming}, 151(1):3--34, 2015.

\bibitem{Xian20arxiv_stnif}
Wenqi Xian, Jia-Bin Huang, Johannes Kopf, and Changil Kim.
\newblock Space-time neural irradiance fields for free-viewpoint video.
\newblock {\em https://arxiv.org/abs/2011.12950}, 2020.

\bibitem{lpips}
Richard Zhang, Phillip Isola, Alexei~A. Efros, Eli Shechtman, and Oliver Wang.
\newblock The unreasonable effectiveness of deep features as a perceptual
  metric.
\newblock {\em 2018 IEEE/CVF Conference on Computer Vision and Pattern
  Recognition}, Jun 2018.

\bibitem{zhou2021cips3d}
Peng Zhou, Lingxi Xie, Bingbing Ni, and Qi Tian.
\newblock Cips-3d: A 3d-aware generator of gans based on
  conditionally-independent pixel synthesis, 2021.

\bibitem{NEURIPS2018_92cc2275}
Jun-Yan Zhu, Zhoutong Zhang, Chengkai Zhang, Jiajun Wu, Antonio Torralba, Josh
  Tenenbaum, and Bill Freeman.
\newblock Visual object networks: Image generation with disentangled 3d
  representations.
\newblock In S. Bengio, H. Wallach, H. Larochelle, K. Grauman, N. Cesa-Bianchi,
  and R. Garnett, editors, {\em Advances in Neural Information Processing
  Systems}, volume~31. Curran Associates, Inc., 2018.

\end{thebibliography}

\end{document}